\documentclass{article}

\usepackage{natbib}
\usepackage{comment}

\usepackage[final]{neurips_2022}

\usepackage[utf8]{inputenc} 
\usepackage[T1]{fontenc}    
\usepackage{hyperref}       
\hypersetup{
    colorlinks = false,
    linkbordercolor = {0 1 0} }
\usepackage{url}            
\usepackage{booktabs}       
\usepackage{amsfonts}       
\usepackage{nicefrac}       
\usepackage{microtype}      
\usepackage{xcolor}         
\usepackage{amsmath}
\usepackage{multirow}
\usepackage{graphicx}
\usepackage{graphics}
\usepackage{wrapfig}
\usepackage{caption} 
\usepackage{subcaption}
\usepackage{longtable}
\usepackage{algorithm}
\usepackage{algpseudocode}
\usepackage{pifont}

\definecolor{myred}{rgb}{0.8,0,0}
\definecolor{mygreen}{rgb}{0,0.6,0}
\definecolor{myblue}{rgb}{0,0,0.7}
\definecolor{mypurple}{rgb}{0.6,0,0.8}
\definecolor{bluegray}{rgb}{0.4, 0.6, 0.8}

\definecolor{deletecolor}{rgb}{0.5,0.5,0.5}

\title{Grounding Large Language Models in Interactive Environments with Online Reinforcement Learning}

\author{
  Thomas Carta$^*$ \\
 Inria (Flowers)\\
 University of Bordeaux, France  \\
  \texttt{thomas.carta@inria.fr} \\
  \And
  Clément Romac$^*$ \\
  Inria (Flowers) \\ 
  University of Bordeaux, France \\
  Hugging Face  \\
  \texttt{clement.romac@inria.fr} \\
\And
  Thomas Wolf \\
  Hugging Face  \\
  \And
  Sylvain Lamprier \\
  Univ Angers, LERIA, \\ SFR MATHSTIC, F-49000 Angers, France\\
  \And
  Olivier Sigaud \\
  Sorbonne Université, ISIR, Paris, France \\
  \And
  Pierre-Yves Oudeyer \\
  Inria (Flowers) \\
  University of Bordeaux, France \\
}

\begin{document}
\definecolor{grey_color}{RGB}{68,68,68}

\maketitle
\def\thefootnote{*}\footnotetext{These authors contributed equally to this work}\def\thefootnote{\arabic{footnote}}

\begin{abstract}
Recent works successfully leveraged Large Language Models' (LLM) abilities to capture abstract knowledge about world's physics to solve decision-making problems. Yet, the alignment between LLMs' knowledge and the environment can be wrong and limit functional competence due to lack of grounding. In this paper, we study an approach (named GLAM) to achieve this alignment through functional grounding: we consider an agent using an LLM as a policy that is progressively updated as the agent interacts with the environment, leveraging online Reinforcement Learning to improve its performance to solve goals. Using an interactive textual environment designed to study higher-level forms of functional grounding, and a set of spatial and navigation tasks, we study several scientific questions: 
1) Can LLMs boost sample efficiency for online learning of various RL tasks?
2) How can it boost different forms of generalization? 3) What is the impact of online learning? 
We study these questions by functionally grounding several variants (size, architecture) of FLAN-T5.
\end{abstract}

\section{Introduction}\label{sec:introduction}
The recent rise of Transformer-based Large Language Models (LLMs) trained on massive text datasets in Natural Language Processing has led to models exhibiting impressive capabilities (e.g. natural language generation, question answering, reasoning, translation...) \citep{devlin-etal-2019-bert, brown-etal-2020-language, rae_scaling_2022, chowdhery_palm_2022, scao_bloom_2022}. Recently, LLMs were shown to capture aspects of the physical rules in our world, e.g. about space \cite{patel_mapping_2021}, colors \cite{abdou2021can} or even affordances between bodies and objects \cite{ahn-etal-2022-do}. This form of prior knowledge was exploited to suggest plans of action to solve goals in robotics \cite{huang-etal-2022-inner, ahn-etal-2022-do, liang-2022-code}. However, LLMs are known to suffer from a lack of grounding which prevents them from properly dealing with the meaning of inter-related concepts and their use for functional competence in interactive environments \cite{mahowald-etal-2023-dissociating}. Indeed, alignment between statistical structures in such LLMs and environments can be very limited, or even sometimes entirely wrong. This is partly due to 1) a training process (predicting next words) that is not directly incentivized to solve problems in an environment, 2) lack of abilities to intervene in the environment to identify causal structures; 3) lack in abilities to learn based on data collected as a result of interacting with the environment \citep{bender_climbing_2020, bisk_experience_2020}.

In the literature, language grounding has referred to various related objectives \cite{thill-etal-2014-on}. First, symbol grounding can be formulated as the general problem of connecting a symbol system \citep{harnad1990symbol}, internal to an agent, to the environment, in such a way that internal processing of these symbols can be used to to act appropriately in this environment. One dimension of this problem is associating "elementary" symbols, such as the names of objects, with invariant structures in high-dimensional perceptual modalities such as vision \cite{cangelosi2010integration, wiriyathammabhum-et-al-2017}. Such a grounding, called "direct grounding", has been extensively studied in the past leading to various efficient methods \citep{alayrac-etal-2022-flamingo, radford-etal-2021-learning, lu_unified-io_2023}, included in the context of robotic bodies \cite{cangelosi_review_2018}. Another dimension is how to ground higher-order symbolic tokens, or abstract concepts, into elementary symbols, often through approaches such as distributional semantics \cite{harris-1954-distributional, boleda-2020-distributional}. This has been called "grounding transfer" \cite{cangelosi_review_2018}. Beyond such mere associations, a key question about grounding is how internal processes that manipulate symbols can model, predict and control external physical and social processes: they need to be aligned on and constrained by these external dynamics and relational structures (at various levels of abstraction). This last notion of grounding, which we refer here as "functional grounding", is relative to a particular environment which may be the human physical environment but also more abstract interactive environments simulated in computers (where abstract physics can differ from human environments).

In this paper, we consider interactive textual worlds \cite{cote-etal-2019-textworld, jansen_systematic_2021}, which are precisely designed to focus on these higher-level forms of functional grounding. In textual worlds, environments can encode rich forms of physical structures inspired by the ones in the human world, e.g. \cite{wang-etal-2022_scienceworld}, yet agents act and perceive in these environments only through the textual modality. In this context, this paper aims to make progress towards the following largely open question:
how could LLMs be used as agent policies producing actions towards goals in interactive environments, perceiving the outcome of these actions, and incrementally grounding
and updating their knowledge with the new observations they collect? 

Building on recent works successfully using Reinforcement Learning (RL) to finetune LLMs for natural language generation tasks \citep{stiennon-etal-2020-learning, ouyang-etal-2022-training, ramamurthy-etal-2022-is}, we propose the first study about functional grounding of LLMs through incremental online RL. In particular, we aim at empirically answering the following open scientific questions:  
\par\smallskip $\bullet$
\textbf{Q1. Sample efficiency} How fast can an LLM adapt and learn to solve various spatial and navigation problems specified in natural language? How does the use of pre-trained knowledge from LLM boosts sample efficiency?

\par\smallskip $\bullet$ \textbf{Q2. Generalization to new objects}: Once functionally grounded, how can an LLM generalize to various kinds of changes about objects, yet staying in trained tasks?

\par\smallskip $\bullet$ \textbf{Q3. Generalization to new tasks}: How can such an interactively trained LLM perform zero-shot generalization to new tasks? How does generalization depend on the kind of new tasks?
\par\smallskip $\bullet$ \textbf{Q4. Impact of online interventions}: What is the empirical impact of grounding using online RL with incremental interactions in comparison with offline Behavioral Cloning from a dataset of expert trajectories?

\begin{figure*}[htb]
    \vskip 0.2in
    \begin{center}
        \centerline{\includegraphics[width=\textwidth]{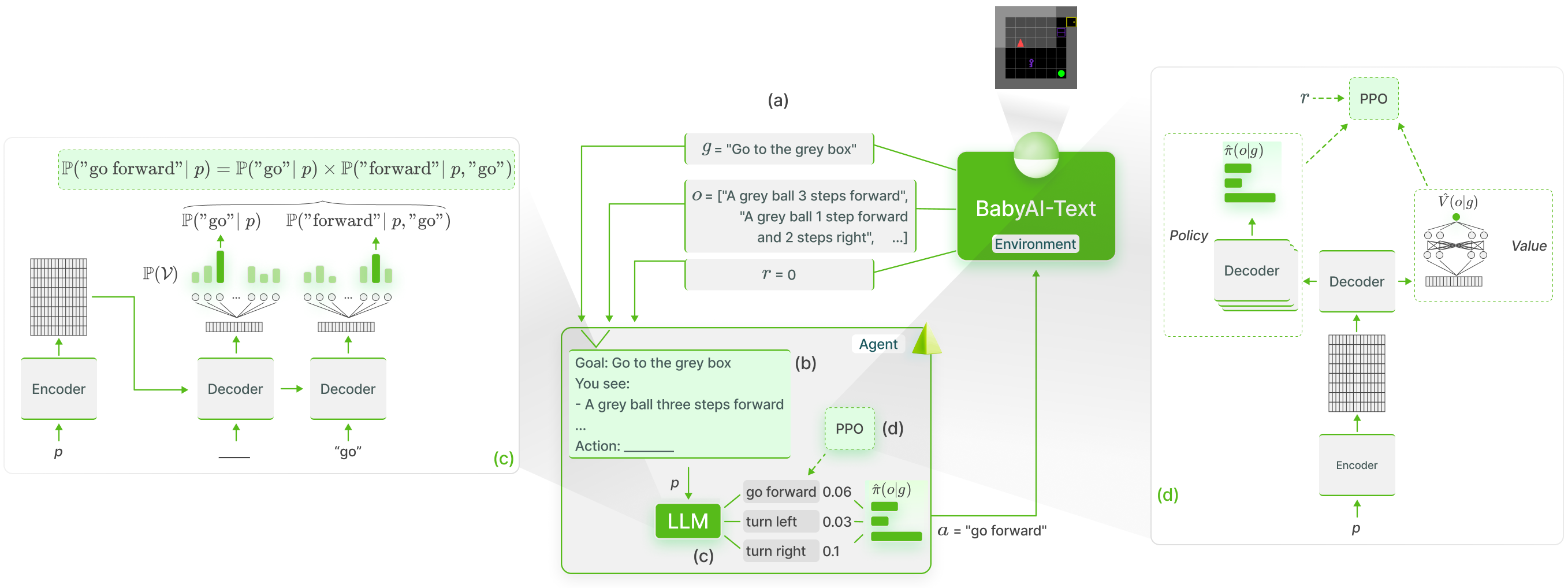}}
        \caption{\textbf{The GLAM method: we use an LLM as agent policy in an interactive textual RL environment (BabyAI-Text) where the LLM is trained to achieve language goals using online RL (PPO), enabling functional grounding}. (a) BabyAI-Text provides a goal description for the current episode as well as a description of the agent observation and a scalar reward for the current step. (b) At each step, we gather the goal description and the observation in a prompt sent to our LLM. (c) For each possible action, we use the encoder to generate a representation of the prompt and compute the conditional probability of tokens composing the action given the prompt. Once the probability of each action is estimated, we compute a softmax function over these probabilities and sample an action according to this distribution. That is, the LLM is our agent policy. (d) We use the reward returned by the environment to finetune the LLM using PPO. For this, we estimate the value of the current observation by adding a value head on top of our LLM. Finally, we backpropagate the gradient through the LLM (and its value head).}
        \label{fig:main_schema}
    \end{center}
    \vskip -0.2in
\end{figure*}

To answer these scientific questions in Section \ref{sec:experiments}, we present a functional grounding method for LLMs (see \figurename~\ref{fig:main_schema} and Section \ref{sec:methods}), and transpose the BabyAI environment \citep{chevalierboisvert-etal-2019-babyai} into a textual version. Additionally, we aim to help the RL community further develop grounding techniques for LLMs in interactive environments by releasing, in addition of the code of this paper\footnote{\url{https://github.com/flowersteam/Grounding_LLMs_with_online_RL}}, a Python library named \textit{Lamorel}\footnote{\url{https://github.com/flowersteam/lamorel}} facilitating the use of LLMs at scale for RL practitioners. While many tools already exist for LLMs and NLP tasks, moving to an RL setting with interactive environments requires adaptations (e.g. very frequent need of fast inference to compute action probabilities) making previous tools not well suited for RL practitioners (see Section \ref{sec:methods_lamorel}).

\section{Related work}
\paragraph{Language-conditioned RL}
We position our work in the Language-conditioned RL setting, where an  \textit{instruction-following} agent learns a policy that executes actions in an interactive environment in order to fulfill a language instruction \citep{luketina2019survey}. While several works studied this setting for various tasks in 2D or 3D environments \citep{hermann-etal-2017-grounded, misra-etal-2017-mapping, bahdanau-etal-2018-learning, colas-etal-2020-language, chevalierboisvert-etal-2019-babyai}, we here focus on text-only interactions (i.e. performing textual commands given textual observations) as in \citet{shridhar-etal-2021-alfword}. However, our work studies how LLMs can not only encode this instruction \citep{hill_human_2020} but also be directly used as agent policies choosing actions given the observation. 

\paragraph{Textual environments for RL}
Many text-only environments have been used and developed \citep{jansen_systematic_2021,wang-etal-2022_scienceworld}. They usually implement high-level text commands along with very large action spaces and complex dynamics between entities, often aiming to study functional grounding of abstract policies. While these environments offer interesting properties, we had to introduce a new one given the purpose and constraints of our study. Dealing here with computationally expensive LLMs, we chose to trade complex action spaces for systematic experiments studying the questions of the introduction. Second, to perform an in-depth analysis of our functional grounding method, we focused on lower-level navigation skills in spatial environments (which lacks in most textual environments as the agent can usually just change room and has direct access to objects in a room). Moreover, several ablation studies shown in Appendix \ref{appendix:additional_results_action_space_distractors} required precise control over the procedural generation (usually not offered by textual environments). For these reasons, we adapted the BabyAI platform \citep{chevalierboisvert-etal-2019-babyai} into a procedural text-only version that enables decoupling exploration challenges from perception challenges. 
Additionally, we are still able to use BabyAI's visualization tools to analyze trajectories (see \figurename~\ref{fig:main_schema}).

\paragraph{Foundation Models for decision making}
Foundation models trained on massive datasets were shown to exhibit impressive abilities along with fast adaptation to a wide range of downstream tasks in vision \citep{yuan-etal-2021-florence}, language \citep{devlin-etal-2019-bert,brown-etal-2020-language} and cross-modalities \citep{ramesh-etal-2021-zero, jiang-etal-2021-vima, alayrac-etal-2022-flamingo}. While such abilities have been leveraged to provide reward to RL agents \cite{gupta-etal-2022-foundation, fan-etal-2022-minedojo}, a recent line of work started focusing on using Foundation Models (and in particular LLMs) to guide agents policy. 

First, SayCan \citep{ahn-etal-2022-do}, Code as Policies \citep{liang-2022-code} and Inner Monologue \citep{huang-etal-2022-inner} used LLMs as high-level planners in robotics setups. Because their LLM is not directly used as agent policy for low-level actions and is not grounded using its interactions with the environment, \citet{ahn-etal-2022-do} had to use an external affordance function to re-rank the actions proposed by the LLM. Similarly, \citet{yao-etal-2022-react} also featured a closed-loop feedback between an LLM that is the planner and an agent that is the actor but this time in a textual environment. Expanding on this, \citet{dasgupta-etal-2022-collaborating} added a reporter observing the environment and reporting useful information to the planner. While hinting at the usefulness of prior knowledge contained in LLMs for embodied tasks, these works are limited by the absence of grounding.

Second, several works proposed to first finetune LLMs on expert trajectories before using them in the environment. Using their ScienceWorld benchmark, \citet{wang-etal-2022_scienceworld} showed that LLMs finetuned using Behavioral Cloning performed worse than a much smaller and randomly initialized Deep Q-Network trained using RL supporting the hypothesis that grounding in the environment through direct interactions is crucial. Finally, \citet{reid-etal-2022-can} reused LLMs to perform offline RL in non-linguistic environments leveraging the internal structures learned by LLMs but no longer using words or symbols they were trained to manipulate (\citet{takagi-2022-on} investigated how these internal structures can be relevant for unrelated tasks).

Finally, one may also pretrain a policy using Behavioral Cloning or offline RL from expert trajectories before finetuning it with interactions with an environment. Related to our work, the Online Decision Transformer \citep{zheng_online_2022} first uses offline RL to pretrain a transformer model and eventually finetunes it with online RL. But compared to our study, it did not use a general Language Modeling pretraining objective and therefore did not study functional grounding of language symbols.


\paragraph{Finetuning LLMs with RL}
Recent works successfully leveraged RL to finetune LLMs. RL was used in particular to improve alignment between generated text and human preferences \citep{stiennon-etal-2020-learning, ouyang-etal-2022-training, ramamurthy-etal-2022-is}. In this Reinforcement Learning from Human Feedback (RLHF) framework, text generation is viewed as a sequential decision-making problem where each "action" of the LLM is a new token and the "state" corresponds to the prompt. Most of these methods used PPO \citep{schulman-etal-2017-proximal} to finetune their LLMs using a reward function learned on a dataset of collected human interactions. With this technique, \citet{ouyang-etal-2022-training} managed to generate more human-aligned outputs despite having a model (InstructGPT) with $100$ times fewer parameters than GPT-3 \citep{brown-etal-2020-language}. While our work shares the PPO-based finetuning with RLHF, our setup diverges from it in multiple aspects. First, our LLM is functionally grounded using an external task-conditioned reward from the environment (which happens to be sparse in our BabyAI-Text environment) and not a learned reward model. Second, the RLHF setup has no external environment dynamics controlling the next state given a previous state and an action (the next state in RLHF is just the previous state with the last generated token appended). In comparison, our work exposes an outer loop controlled by the environment whose dynamics, providing the next state and reward, are unknown to the LLM (in comparison to RLHF where the RL loop is an inner loop in the token generation process). We believe using RL to finetune LLMs can be taken from a broader perspective in which both our framework and RLHF are particular applications.

\section{GLAM: Grounding LLMs with online RL} \label{sec:methods}
We introduce the GLAM method (for Grounded LAnguage Models) where an LLM is used as agent policy and is functionally grounded in an interactive environment using online RL, leveraging collected observations and rewards to improve itself towards achieving goals formulated in language.
We detail this method in the following paragraphs and redirect the reader to \figurename~\ref{fig:main_schema} for a schematic view. We first formalize the textual RL problem we tackle (a). Then, we detail how we use an LLM as agent policy to interact with BabyAI-Text (b, c). Finally, we explain how online RL finetuning is used to ground the LLM in BabyAI-Text (d).

\subsection{Problem statement} \label{sec:methods_textual-rl}
We assume a textual RL setting where, given a language vocabulary $\mathcal{V}$, our environment returns an observation $o \in \mathcal{V}^N$ and a reward $r \in \mathbb{R}$ following an action $a \in \mathcal{A} \subset \mathcal{V}^N$ (i.e. actions are sequences of tokens). We also assume a task or goal description $g \in \mathcal{G} \subset \mathcal{V}^N$ which conditions the reward. Such an environment can be framed as a goal-augmented Partially Observable Markov Decision Process $\mathcal{M = (S,V,A,T,R,G,O,\gamma)}$ with $\mathcal{S}$ the state space, $\mathcal{A} \subset \mathcal{V}^N$ the action space, $\mathcal{G} \subset \mathcal{V}^N$ the goal space, $\mathcal{T: S \times A \mapsto S}$ the transition function, $\mathcal{R : S \times A \times G \mapsto \mathbb{R}}$ the goal-conditioned reward function, $\mathcal{O: S \mapsto V}^N$ the observation function mapping a state to a textual description and finally $\gamma$ the discount factor.

In this work, we extend the BabyAI platform \citep{chevalierboisvert-etal-2019-babyai} initially designed for grounded language learning and propose a text-only extension named BabyAI-Text. We leverage BabyAI's inner procedurally generated minigrid environment where an agent navigates and interacts with objects through $6$ text commands: \textit{turn left, turn right, go forward, pick up, drop and toggle}. We also reuse the set of tasks introduced in BabyAI as well as their associated description along with the sparse scalar reward. Our key difference is the textual description $o \in \mathcal{V}^N$ of the agent's partial observation returned by BabyAI-Text instead of the symbolic representation initially returned by BabyAI (see Appendix \ref{appendix:babyai_text}). We leverage BabyAI-Text in Section \ref{sec:experiments} to assess our grounding method. 

\subsection{LLMs as policies in interactive environments} \label{sec:methods_interact}
In order to use the LLM as the policy in such a textual interactive environment, we gather the task description, the textual description of the current observation and the set of possible actions in a prompt used to feed the LLM. We chose a single arbitrary and simple prompt template (see Appendix \ref{appendix:prompt_examples} for examples) and did not perform any intensive prompt engineering. Indeed, as we finetune the LLM, we expect it to adapt to the chosen prompt template. Nonetheless, a more careful design of prompts could improve the results shown in Section~\ref{sec:experiments}.

Given this prompt, we now need the LLM to output a probability distribution over the possible actions $\mathbb{P}(\mathcal{A})$. For this, \citet{huang-etal-2022-languages, li-etal-2022-pretrained, wang-etal-2022_scienceworld} used the LLM to generate text. If the generated sequence of characters corresponds to one of the possible actions (i.e. $s \in \mathcal{A}$), this action is chosen by the agent. Otherwise, an ad-hoc mapping must be performed to select an action $a_i \in \mathcal{A}$ given $s$. As an alternative method, one could also use more standard RL practices by adding action heads - a Multi-Layer Perceptron (MLP) with $|\mathcal{A}|$ outputs - on top of the LLM. Finally, \citet{ahn-etal-2022-do} proposed to directly use the LLM to compute the (log) probability of each action $a_i \in \mathcal{A}$ by computing the conditional probability of each token in action $a_i = \{w_0, ..., w_{|a_i|}\}$ given the prompt $p$:

\begin{equation}
\label{eq:proba_llm}
    \mathbb{P}_{LLM}(a_i|p) = \prod^{|a_i|}_{j=0} \mathbb{P}_{LLM}(w_j|p, w_{<j})
\end{equation}

with $\mathbb{P}_{LLM}(w_j|p, w_{<j})$ the probability computed by the LLM of token $w_j$ given prompt $p$ and previous tokens $w_{<j}$ (see (c) from \figurename~\ref{fig:main_schema}). This method suffers from requiring a forward pass on the LLM for each action to compute the probability of its sequence of tokens (especially in comparison to new action heads that require a single forward pass on the prompt to compute all actions' probability). However, it also has several advantages, in particular, 1) there is no need of potential ad-hoc mapping as when text is generated, 2) we use only pretrained operations from the LLM and leverage language modeling heads' prior and 3) this method is robust to any action space and can thus be used on any textual environment with no change.

For these reasons, we chose the latter method. We first use log probabilities instead of normalized probabilities using $\mathbb{LP}_{LLM}(a_i|p) = \sum^{|a_i|}_{j=0} log\ \mathbb{P}_{LLM}(w_j|p, w_{<j})$ in replacement of $\mathbb{P}_{LLM}(a_i|p)$ to avoid multiple normalization operations. We eventually normalize all the log probabilities to obtain a distribution over $\mathcal{A}$ using a softmax function:
\begin{equation}
\label{eq:proba_action_space}
    \mathbb{P}(a_i|p) = \frac{e^{\mathbb{LP}_{LLM}(a_i|p)}}{\sum_{a_j \in \mathcal{A}} e^{\mathbb{P}_{LLM}(a_j|p)}}.
\end{equation}

\subsection{PPO finetuning}
We now propose to leverage experiences gathered by the LLM to perform functional grounding. More formally, we aim to learn a policy $\pi : O \times \mathcal{G} \mapsto \mathbb{P}(\mathcal{A})$ that maximizes the expected discounted sum of rewards for any given goal $g \in \mathcal{G}$. We use for this the PPO algorithm \citep{schulman-etal-2017-proximal} that both learns a policy $\hat{\pi} : O \times G \mapsto \mathbb{P}(A)$ and a value function $\hat{V} : \mathcal{O} \times \mathcal{G} \mapsto \mathbb{R}$ approximating the true value $V(s,g) = \mathbb{E}_{a \sim \hat{\pi}(\mathcal{O}(s),g)}\bigl[R(s,g,a) + \gamma V(\mathcal{T}(s,a),g)\bigr]$. 

As mentioned in Section \ref{sec:methods_interact}, we compute the probability of each action $a_i \in \mathcal{A}$ using the likelihood computed by the LLM as $\hat{\pi}(a_i|o,g) = \mathbb{P}(a_i|p)$.

For value approximation, we we add an MLP with a single output for the value on top of the last layer of the first Decoder block (i.e.\ in place of the language modeling heads) in order to compute $\hat{V}(o|g) = \hat{V}(p)$ (see (d) from \figurename~\ref{fig:main_schema}). This position is explained by the fact that we use Encoder-Decoder LLMs in our experiments but our method could easily be used with Decoder-only models by attaching the value head to the Decoder block encoding the last token of the prompt. 

\subsection{Distributed LLM policies using \textit{Lamorel}} \label{sec:methods_lamorel}
Using LLMs to compute probabilities over action space is computationally expensive as it requires computing $\prod^{|a_i|}_{j=0} \mathbb{P}_{LLM}(w_j|p, w_{<j})$ for each action $a_i = \{w_0, ..., w_{|a_i|}\}$. When one uses very large LLMs (i.e. more than hundreds of million parameters), computing the probability of a single action already means performing a long forward pass over the whole network. As a result, computing the probability of each possible action at every step becomes very slow. Considering the number of interactions usually required to solve tasks in BabyAI (and by extension BabyAI-Text), performing online RL finetuning of LLMs easily became intractable with a single LLM distributed over multiple GPUs. To overcome this, we deployed $N$ LLM workers each handling a subset of actions to score in parallel (allowing a quasi-linear time decrease with $N$). We add to this distributed inference the possibility to also perform distributed training (i.e. compute the gradient of minibatches in parallel and gather gradients before updating models). We wrap all this in a Python library named \textit{Lamorel} designed for RL practitioners eager to use LLMs. It allows one to use LLMs as black-box but also to perform more advanced methods such as adding new heads on top of them. See Appendix \ref{appendix:distributed} for more details.

\section{Experiments}\label{sec:experiments}
We design a set of experiments in BabyAI-Text aiming to provide answers for the scientific questions introduced in Section \ref{sec:introduction}. In these experiments, we use Flan-T5 780M \citep{rae_scaling_2022} for 1) the close link between its training corpus (containing instruction-following documents) and our language-conditioned interactive environment, and 2) its simple open-source access through the Hugging Face tools\footnote{\url{https://huggingface.co/docs/transformers/model_doc/flan-t5}}. We apply our GLAM method to Flan-T5 (which we name GFlan-T5 in experiments below for Grounded Flan-T5) and compare it with three baselines. First, we also train a non-pretrained Flan-T5 where we only reuse the pretrained embedding layer and add action heads on top of it (see \figurename~\ref{fig:NPAE-Flan-T5} in appendices). As for GFlan-T5, we propagate the gradient through the entire graph (included the action heads here). We call this baseline NPAE-Flan-T5 (Non-Pretrained with Action heads and Embedding Flan-T5). We show in Appendix~\ref{appendix:additional_results_impact_pretraining} that using a non-pretrained Flan-T5 while keeping the scoring method fails. We also provide as a more classic RL baseline a DRRN \citep{he_deep_2016} agent of approximately $1$M parameters which is often used for TextWorlds. We especially reuse the implementation from \citet{wang-etal-2022_scienceworld} which gave SOTA results and outperformed LLMs. At each step, we feed our $3$ agents above with the following prompt template filled using the information returned by BabyAI-Text (see Appendix~\ref{appendix:prompt_examples} for examples):
\par\smallskip $\bullet$ A header listing what actions are accessible (but not necessarily useful) in the environment in the form of: \\
    \textit{Possible action of the agent: $<$list of actions$>$}
\par\smallskip $\bullet$ The goal of the agent:
    \textit{Goal of the agent: $<$goal$>$}
\par\smallskip $\bullet$ The $3$ previous observations and last $2$ actions, used as a short-term memory required to complete BabyAI-Text tasks (in comparison, the DRRN uses recurrent layers to deal with short-term memory requirements): \\
    \textit{Obs. 0: $<$description from BabyAI-Text at step $t-2>$} \\
    \textit{Action 0:$<$action chosen by the agent at step $t-2>$} \\
    \textit{Obs. 1: $<$description from BabyAI-Text at step $t-1>$} \\
    \textit{Action 1: $<$action chosen by the agent at step $t-1>$} \\
    \textit{Obs. 2: $<$description from BabyAI-Text at step $t>$} \\
    \textit{Action 2: $<$the next action to be chosen by the agent$>$}

Finally, as BabyAI-Text simply provides an alternative mapping of observations, we add as an indication the performance of the PPO agent used in \citep{chevalierboisvert-etal-2019-babyai} that runs on BabyAI rather than BabyAI-Text (i.e. using symbolic observations instead of textual descriptions) and name this agent Symbolic-PPO in results below. In Appendix~\ref{appendix:additional_results_representation_impact}, we show that symbolic observations provided by BabyAI encode biases that ease learning compared to text descriptions. However, even with this advantage, GFlan-T5 outperforms Symbolic-PPO in all our setups.

We first study Q1 by training the different agents in a multi-task setting assessing their efficiency at learning the different tasks. We then address questions Q2, Q3 and Q4 using a set of generalization experiments (\figurename~\ref{fig:full_gene}) on the zero-shot abilities of the resulted trained agents mostly inspired from \citep{colas-etal-2020-language} and \citep{valmeekam-etal-2022-large}. We report their average success rate as well as standard deviation. We compare the results of GFlan-T5, DRRN as well as Flan-T5 (i.e. the LLM used in GFlan-T5 but before our finetuning) to show how our grounding method impacted it. All results below are given with their $99\%$ confidence interval (mathematical details are given in Appendix \ref{appendix:confidence_interval}).

\subsection{How fast can an LLM adapt and learn to solve tasks? (Q1)}
\label{sec:results_q1}
In order to study question Q1, we train our agents for $1.5$ million steps in BabyAI-Text where each episode is a task randomly sampled from the following: 
\begin{itemize}
    \item \textbf{Go to $<$object$>$}, a simple navigation task that requires reasoning abilities to choose the right plan given objects' position;
    \item \textbf{Pick up $<$object$>$}, a reasoning task that combines navigation tasks;
    \item \textbf{Put $<$object A$>$ next to $<$object B$>$}, which requires first reaching $<$object A$>$, picking it up, reaching $<$object B$>$ and finally dropping $<$object A$>$ next to $<$object B$>$;
    \item \textbf{Pick up $<$object A$>$ then go to $<$object B$>$} and \textbf{Go to $<$object B$>$ after pick up $<$object A$>$}, both serving to test reasoning abilities on temporal sequences;
    \item \textbf{Unlock $<$door$>$}, a task that includes inferring that a key is needed to unlock the door, finding the right key (i.e. the one colored as the door) and eventually using the toggle action with the key on the door.
\end{itemize}    
In each task, the agent must navigate in one procedurally generated room with $8$ distractors (i.e. useless objects for the completion of the task).

\begin{figure}[ht]
    \vskip 0.2in
    \begin{center}
        \centerline{\includegraphics[width=0.7\columnwidth]{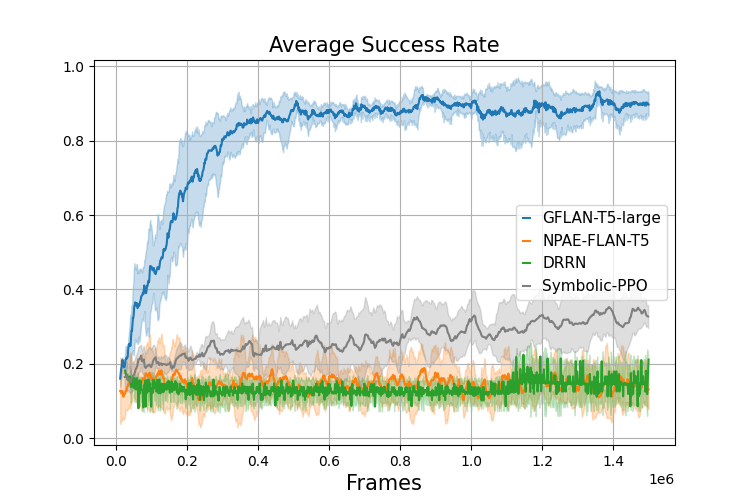}}
        \caption{\textbf{Q1. Sample efficiency}: Evolution over $4$ seeds of the average success rate and standard deviation on all Q1 tasks. The details of average success rate calculation over the goals is given in Appendix~\ref{appendix:average_sr_over_goals}}
        \label{fig:mtrl}
    \end{center}
    \vskip -0.2in
\end{figure}

We plot the mean and standard deviation of the success rate (i.e. $1$ if the goal has been reached, $0$ otherwise) over $4$ seeds of GFlan-T5, NPAE-Flan-T5, DRRN and Symbolic-PPO in \figurename~\ref{fig:mtrl}. In addition, we also monitor the evolution of probability of each possible action on a set of $11$ evaluation prompts to assess agents' abilities to solve each task in Appendix \ref{appendix:prompt_examples}. By plotting the evolution of the distribution over possible actions in \figurename~\ref{fig:policy_evolution}, we better grasp how and when the agents learn skills (e.g. navigation skills).

Looking at the evolution of the average success rate, GFlan-T5 quickly reaches $0.8$ after only $250.000$ steps (and $0.9$ after approximately $600.000$ steps). In comparison, both DRRN and NPAE-Flan-T5 are still under $0.2$ after $1.5$ million steps. Even when compared to Symbolic-PPO, which uses symbolic observations (easier to process than language as shown in Appendix~\ref{fig:compare_gtrb_drrn_ppo}), GFlan-T5 exhibits a drastically better sample efficiency with Symbolic-PPO almost reaching $0.4$ after $1.5$ million steps. \figurename~\ref{fig:policy_evolution} and Table~\ref{tab:prompt} highlight how GFlan-T5 leverages its knowledge about the relationships between entities to learn navigation tasks in less than a hundred updates.

The failure of NPAE-Flan-T5 both highlights how GFlan-T5 leverages the LLM's pretrained knowledge to deal with the proposed tasks and how the finetuning method helps achieve the grounding objective. Furthermore, the fact that GFlan-T5 strongly outperforms Symbolic-PPO and the latter is better than NPAE demonstrates how language can be used as a tool to scaffold learning if already acquired. It also exaplins how counterproductive it can be if one asks an agent to both learn a task and language at the same time (see Appendix~\ref{appendix:additional_results_representation_impact} for further results). 

We now perform a deeper analysis of this sample efficiency by studying the impact of varying the action space and the number of distractors. We provide both the evolution of success rate and a sample efficiency measure $SE$:
\begin{equation}
    \label{eq:se}
    SE = \frac{1}{T} \sum_{t=0}^{T} SR_t
\end{equation}
where $T$ is the number of steps or frames seen and $SR$ the success rate at frame $t$.

\begin{figure}
\centering
\begin{minipage}[t]{.47\textwidth}
    \centering
    \includegraphics[width=\linewidth]{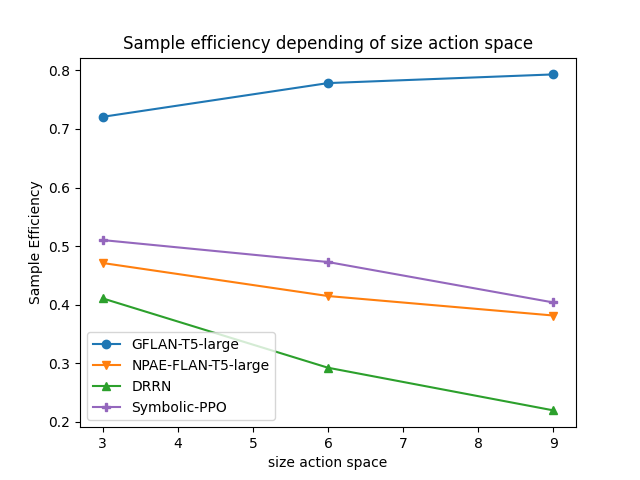}
    \subcaption{Impact of the action space size ($3$, $6$ or $9$ actions with always only $3$ useful actions).}
    \label{fig:ablation_se_action_space}
\end{minipage} 
\hfill
\begin{minipage}[t]{.47\textwidth}
    \centering
    \includegraphics[width=\linewidth]{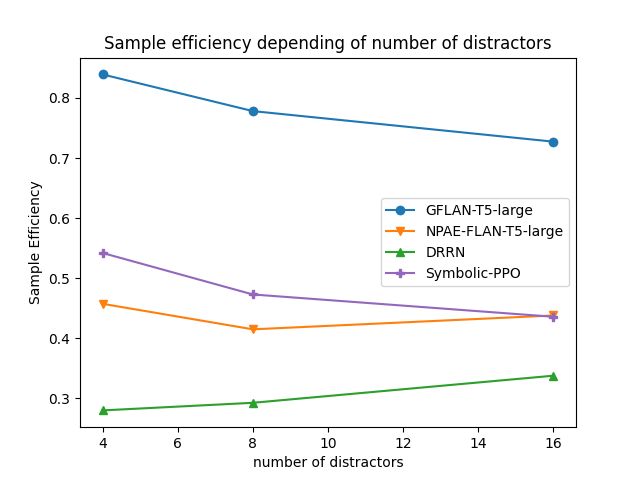}
    \subcaption{Impact of the number of distractors.}
    \label{fig:ablation_se_nbr_dist}
\end{minipage}
\caption{Impact of the aciton space size and number of distractors on the sample efficiency measure (Equation \eqref{eq:se}). We report results averaged over $2$ seeds for training on the \textit{Go To} task.}
\end{figure}

\subsubsection{Impact of the dimension of the action space}
In this experiment, we test the sensitivity of LLMs to the size of the action space by using $3$ different action spaces when trained on the \textit{Go to $<$object$>$} task:
\begin{itemize}
    \item The \textbf{restricted} action space composed of the only $3$ useful actions: \texttt{turn left}, \texttt{turn right}, \texttt{go forward}.
    \item The \textbf{canonical} action space composed of the $6$ actions that can be performed in the environment with $3$ useful and $3$ useless actions that are \texttt{pick up}, \texttt{drop} and, \texttt{toggle} (they are useless here as the agent is only navigating).
    \item The \textbf{augmented} action space composed of $9$ actions ($3$ useful and $6$ useless with \texttt{pick up}, \texttt{drop}, \texttt{toggle}, \texttt{sleep}, \texttt{do nothing} and \texttt{think}). The last three actions have been chosen such that they clearly have no use for the \textit{Go To $<$object$>$} task and consequently should not impact an agent that has knowledge about the world. 
\end{itemize}

We conduct our tests in an environment with $1$ room, $8$ distractors and report results in \figurename~\ref{fig:ablation_se_action_space}. Results show no impact on GFlan-T5, while the performance of other agents decreases with larger action spaces. We hypothesize that this is due to the LLM's ability to discard useless actions quickly at the beginning of finetuning. 

\subsubsection{Impact of the number of distractors}
Similarly, we expect LLMs to be less sensitive to variations in task complexity. We assess this by plotting the evolution of sample efficiency (Equation~\eqref{eq:se}) for $4$, $8$ and $16$ distractors. We conduct these tests in an environment with $1$ room and observe a slight performance loss from GFlan-T5 when the number of distractors increases (\figurename~\ref{fig:ablation_se_nbr_dist}). In comparison, Symbolic-PPO degrades as the number of distractors increases with a success rate decreasing by $38\%$ from $4$ to $16$ distractors whereas the GFlan-T5 success rate only decreases by $14\%$. We hypothesize that the LLM manages to focus on the relevant aspect of the environment quickly.

Thus, GFlan-T5 seems robust with similar learning curves when one increases the action space size (from $3$ to $9$ actions with only $3$ useful ones) or the number of distractors (from $4$ to $16$) We also provide in Appendix \ref{appendix:additional_results} an additional ablation analyzing the impact of the LLM's size~\ref{appendix:additional_results_llm_size}. Results highlight that the number of parameters has a high impact on the learning process. Indeed, we observe a strong difference on sample efficiency and asymptotic performance between a small LLM ($80$ million parameters) and the $780$ million parameters we used here. We also plot the full learning curves for the ablation on the action space size and the number of distractors in appendices \ref{appendix:additional_results_action_space} and \ref{appendix:additional_results_distractors} respectively. 

\begin{figure*}[htb]
    \vskip 0.2in
    \begin{center}
        \centerline{\includegraphics[width=\textwidth]{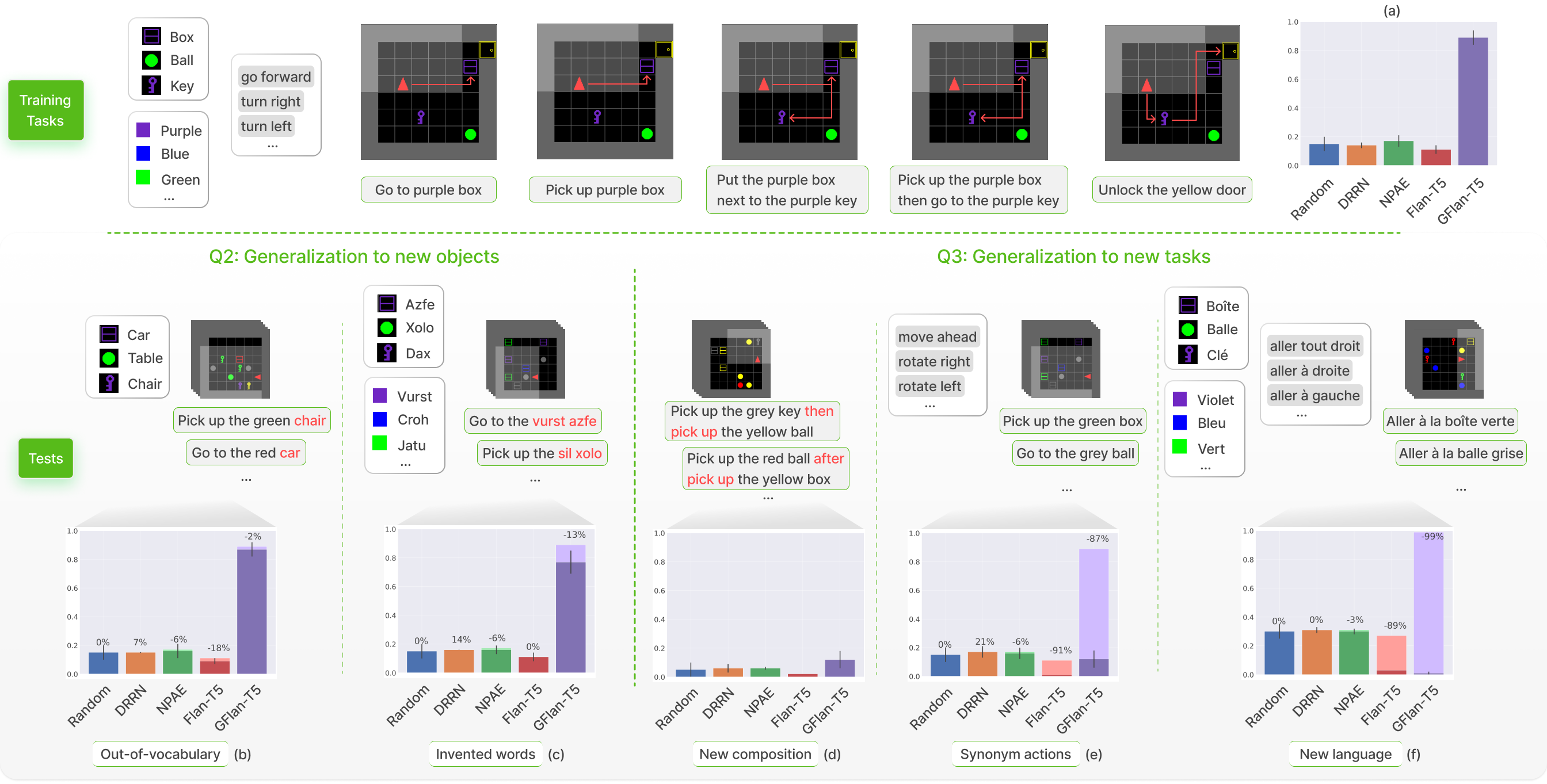}}
        \caption{\textbf{Generalization tests}: We train all agents on a mix of $5$ different tasks and evaluate their generalization abilities on $1000$ test episodes (also containing a mix of these $5$ tasks) (a). We compare them to two baselines: an agent choosing actions randomly (Random) and the zero-shot Flan-T5 (without any finetuning). We then perform several generalization studies to answer Q2 and Q3 by (b) substituting object names out-of-vocabulary names, (c) substituting objects and colors by invented words, (d) testing a new composition of tasks, (e) substituting actions by synonyms and (f) translating the whole environment to French for the \textit{Go To} task. For each agent, we plot its mean success rate over $2$ seeds along with the confidence interval and the delta with performance on the same task without any change (except for (d), on which no baseline result can be provided as this task is completely new). The details of average success rate calculation over the goals is given in Appendix~\ref{appendix:average_sr_over_goals}.}
        \label{fig:full_gene}
    \end{center}
    \vskip -0.2in
\end{figure*}

\subsection{Q2. Generalization to new objects} \label{sec:results_q2} 
In this section, we analyze how a functionally grounded agent can generalize its skills to new objects. Indeed, we expect our agents to focus on the the geometry of the environment (how objects are positioned and how their positioning is described), but not on the identity of the objects themselves (e.g. \textit{Go to $<$object$>$} should be achieved even if the object has not been seen during training). We test if this property is present in our trained agents by measuring their zero-shot performance in two environments. First, an environment with nouns not in the training vocabulary (e.g. "tree")\footnote{The out-of-vocabulary nouns are given in Appendix~\ref{appendix:out_voc}.} and second, an environment with invented objects (made of an invented adjectives and an invented nouns such as \textit{faze dax}).\footnote{The invented objects are given in Appendix~\ref{appendix:invented_voc}.} We use the environment the agent has been finetuned on (i.e. without any word substitutions) as a control environment. Results in \figurename~\ref{fig:full_gene} (Q2 part) indicate that GFlan-T5 is not affected when tasks contain out-of-vocabulary nouns. Moreover, even if the GFlan-T5's success rate decreases by $13\%$ when it is in an environment with invented objects, it still retains strong performances compared to baselines. These results support the hypothesis that GFlan-T5 has functionally grounded the symbols that describe the geometry of the environment and the instructions (e.g. words such as "in front", or the meaning of "steps" as a distance measure)\footnote{See Section~\ref{appendix:babyai_text} for more details on the geometry.}.

\subsection{Q3. Generalization to new tasks} \label{sec:results_q3}
In this Section, we perform generalization tests as in Section \ref{sec:results_q2}, but with new unseen tasks. Using these, we verify to what extent an agent is able to compose and generalize over the symbols it has grounded during finetuning.

\begin{table*}[hbt]
    \caption{Generalization tests for Behavioral Cloning}
    \label{tab:tab_bc}
    \centering
    \begin{tabular}{ll|llll}
        \toprule
        \multicolumn{2}{c}{\textbf{Environments}} & GFlan-T5 & BC-GFlan-T5 & BC-Bot & Random \\
        \midrule
        \multirow{2}{*}{\textbf{Q4}}\hypertarget{tab_q4} & Go To task no change & $0.82 \pm 0.02$ & $0.69 \pm 0.08$ & $0.73 \pm 0.07$ & \multirow{2}{*}{$0.30 \pm 0.05$} \\
        & Go To task with invented words & $0.74 \pm 0.004$ & $0.7 \pm 0.07$ & $0.63 \pm 0.08$ & \\
        \bottomrule
    \end{tabular}
\end{table*}

\paragraph{New composition of learned tasks: Pick up $<$object A$>$ then/after pick up $<$object B$>$}
During finetuning, agents learn to do both 1) \textit{Pick up $<$object A$>$} and 2) \textit{Pick up $<$object A$>$ then go to $<$object B$>$} or \textit{Go to $<$object B$>$ after pick up $<$object A$>$} tasks. We test in this experiment if an agent can compose grounded symbols to solve the new tasks \textit{Pick up $<$object A$>$ $<$then/after$>$ pick up $<$object B$>$}. Results in \figurename~\ref{fig:full_gene} (Q3 part) hint that, while all agents fail to solve these new tasks, GFlan-T5 outperforms other baselines by reaching an $0.12$ success rate compared to Flan-T5 ($0.07$) or Random ($0.05$). These low results can be explained by the fact that none of the agents managed to master the \textit{Pick up $<$object A$>$ then go to $<$object B$>$} or \textit{Go to $<$object B$>$ after pick up $<$object A$>$} tasks during training (see Appendix~\ref{appendix:prompt_examples}). More details about the grounding of "then" and "after" are given in the Appendix~\ref{appendix:tab_groud_after_then}.

\paragraph{Seen tasks with synonym actions} In this task, we test the robustness of our agents to actions by replacing the actions used during training by synonyms. For instance, "go forward" is replaced with "move ahead"\footnote{A table giving all the used synonyms is given in Appendix~\ref{appendix:synonym_voc}}. We expect LLMs, which already learned to map words to an embedding space, to also ground synonyms as they ground words of the environment. In this environment (see \figurename~\ref{fig:full_gene} Q3 part), the success rate of GFlan-T5 is  $0.12$ vs $0.01$ for Flan-T5. Thus the grounding of some words (here the actions) also improves the grounding of their synonyms. However, we observe an $87\%$ drop in performance compared to the original settings, which we assume is due to an over-fitting of the actions' vocabulary. 

\paragraph{New language} In order to understand how far agents can generalize, we test them with a language not seen during training (French). Knowing that Flan-T5 has been pretrained with a multilingual corpus and is able to translate simple sentences, we test whether grounding in GFlan-T5 has also impacted its manipulation of other languages. However, we observe that even only for a simple navigation task (i.e. \textit{Go To}), the model fails to generalize to a new language with a success rate ($0.02$) worse than random ($0.30$). We hypothesize that when too many grounded symbols are modified at once, functional grounding fails to be transferred to this new subsystem of symbols. Complementary experiments that confirm and reinforce this result are presented in appendices~\ref{appendix:generalization_test_q2} and ~\ref{appendix:generalization_test_q3}.

\subsection{What is the impact of using RL vs Behavioral Cloning for grounding? (Q4)}
\label{sec:results_q4}
Finally, we study how online interactions with an environment, enabling learning through interventions and trial-and-error, improves grounding in comparison to pure Behavioral Cloning (BC). We compare a GFlan-T5 trained on the \textbf{Go To} task over $400000$ steps with two baselines trained with Behavioral Cloning using $400000$ transitions (see Appendix~\ref{appendix:bc}). For the baseline called BC-GFlan-T5, transitions are collected from GFlan-T5 finetuned on the \textbf{Go To} task. For BC-Bot, transitions are collected using the BabyAI procedural bot achieving a success rate of $1$.

In Table~\ref{tab:tab_bc}, we measure the success rate of GFlan-T5 and the baselines on two tasks: \textit{Go To} and \textit{Go To} with invented nouns and adjectives. First, once can see that GFlan-T5 outperforms all baselines in both tasks. 
Second, as GFlan-T5 does not achieve a success rate of $1$ on the \textbf{Go To} task, its collected trajectories for BC can contain deceptive transitions in comparison to the ones collected by the bot. Hence, we obtain the expected result that BC-Bot outperforms BC-GFlan-T5. Finally, we expect our agents not to be affected by an environment where nouns and adjectives are replaced by invented ones in such navigation tasks. Experiments show that GFlan-T5 is less affected ($0.82\rightarrow 0.74$) than the BC-Bot ($0.73\rightarrow 0.63$). GFlan-T5 also performs better in the \textit{invented words} task than the BC-GFlan-T5 (success rate of $0.7$). 


\section{Conclusion}
In this paper, we proposed the GLAM method for functional grounding (i.e. aligning internal symbols to external dynamics so that the agent can use them to solve tasks in the environment) of LLMs in interactive textual environments based on online RL. Using our new BabyAI-Text environment, we performed several experiments studying $4$ scientific questions. We showed how GLAM, which requires almost no environment-specific modifications on the LLM, enables to drastically improve performances to solve RL tasks in this environment as compared to zero-shot use the LLM, to supervised finetuning and to RL finetuning of non-pretrained LLMs. We showed how it boosts both sample efficiency and generalization abilities in zero-shot tests (both to new objects and several new tasks). In addition to these key results, we provided in-depth ablations showing the effect of several parameters (e.g. size) on grounding. We believe this method can act as a milestone towards grounding and using LLMs in interaction with our world. However, this study still suffers several limitations, in particular the fact that current experiments are limited to a textual environment, and the computational inefficiency when scaling up the action space and the size of the LLM. This computational inefficiency constrained this paper to using a single environment and rather small LLMs. Yet, improving computational efficiency (or access to more computational resources) could enable to leverage recent multi-modal Foundation models \citep{alayrac-etal-2022-flamingo}) for grounding LLMs in broader environments (e.g. to robotics setups \citep{lu_aw-opt_2021, ahn-etal-2022-do}). Parallel to this, a future direction would be to study how functionally grounding an LLM on a specific environment affects its zero-shot abilities but also its plasticity and ability to acquire new skills in other environments. Moreover, these results hint that using LLMs as agent policies opens an avenue for escaping the \textit{Tabula-Rasa} RL setting and creating much more sample efficient RL agents.

Finally, the recent rise of real-world deployed applications using LLMs highlighted the various societal and ethical challenges of using such models in real-world scenarios. Similar to RLHF, our work studies how to better align LLMs (but this time to environments in which tasks must be solved). While our approach stands as a first important building block for future works making LLMs more in line with their environment, it is not designed to be ready for real-world deployment and thus we do not recommend to use it in such an applicative context.




\begin{ack}
Experiments presented in this paper were carried out using the HPC resources of IDRIS under the allocation 2022-[A0131011996] made by GENCI. This work also has received funding from the European Commission's Horizon Europe Framework Programme under grant agreement No 101070381 (PILLAR-robots project). We would also like to thank Victor Gondat for his kind help on schemas.
\end{ack}

\medskip

\typeout{}
\bibliography{biblio}

\newpage
\appendix

\section*{Appendices}
This supplementary material provides additional results and discussion, as well as implementation details.

\begin{itemize}
    \item Section~\ref{appendix:environments} presents the BabyAI and BabyAI-Text environments. 
    \item Section~\ref{appendix:additional_results}, contains several additional results. We report the per-task success rate at the end of the training (\ref{appendix:success_per_task}) and how we have averaged the results over the tasks in \ref{appendix:average_sr_over_goals}. We also analyze the influence of the observation's structure (i.e. either a symbolic image for the Symbolic-PPO agent or text for LLM based agents) in \ref{appendix:additional_results_representation_impact}. We then study the influence of pretraining in \ref{appendix:additional_results_impact_pretraining} and conduct several ablation tests to understand the influence of the size of the LLM (\ref{appendix:additional_results_llm_size}), the impact of the size of the action space (\ref{appendix:additional_results_action_space}), and the effect of the number of distractors (\ref{appendix:additional_results_distractors}). Eventually, we verify in \ref{appendix:arobustness_domaine_specific_vocabulary} the robustness of our method to domain-specific vocabulary.
    \item Section~\ref{appendix:prompt_examples} is a qualitative analysis of GFlan-T5 during its training on the environment with a mix of tasks. We plot the evolution of the distribution of actions during training for $11$ prompts.
    \item In Section~\ref{appendix:generalization_test}, we detail complementary tests for questions Q2 (\ref{appendix:generalization_test_q2}) and Q3 (\ref{appendix:generalization_test_q3}). We also analyze the functional grounding of temporal symbols "then" and "after" (\ref{appendix:tab_groud_after_then}).
    \item Section~\ref{appendix:distributed} gives details related to the distributed experimental setup.
    \item Section~\ref{appendix:finetuning} reports hyperparameters and implementation details used to finetune the models using PPO or Behavioral Cloning.
    \item In Section~\ref{appendix:confidence_interval}, we detail how the confidence intervals given in \figurename~\ref{fig:full_gene} and Appendix~\ref{appendix:generalization_test} are obtained.
    \item In Section~\ref{appendix:generalization_dicts}, we give the word substitutions used in our generalization experiments from sections \ref{sec:results_q2} and \ref{sec:results_q3}. 
\end{itemize}

\section{Environments} 
\label{appendix:environments}
\begin{figure}[ht]
    \vskip 0.2in
    \begin{center}
        \centerline{\includegraphics[width=\textwidth]{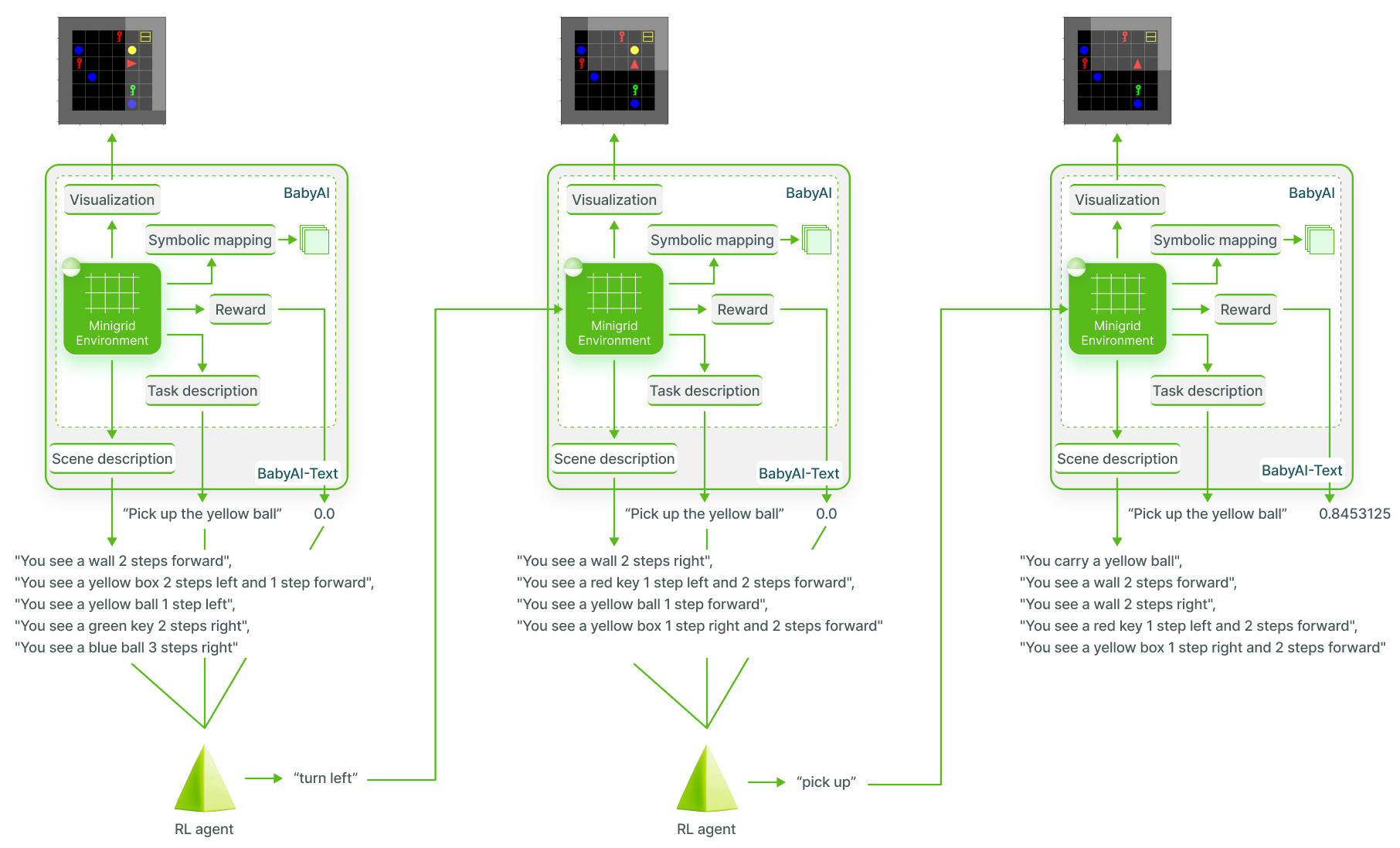}}
        \caption{An illustration of how our BabyAI-Text environment encapsulates BabyAI. We keep the inner minigrid environment as well as task descriptions and reward but map the partial view of the agent to a text description.}
        \label{fig:babyai-text_schema}
    \end{center}
    \vskip -0.2in
\end{figure}

We extend the BabyAI platform \citep{chevalierboisvert-etal-2019-babyai} and create a text-only version named BabyAI-Text that encapsulates BabyAI and returns linguistic observations. \figurename~\ref{fig:babyai-text_schema} explains our environment.

\subsection{BabyAI}
\label{appendix:babyai}
BabyAI \cite{chevalierboisvert-etal-2019-babyai} is a language-conditioned environment where the agent has a limited number of steps to complete a language goal. This platform relies on a gridworld environment (MiniGrid) to generate a set of complex instructions-following environments. It has been specifically designed for research on grounded language learning and related sample efficiency problems. The gridworld environment is populated with the agent and objects (of $6$ possible colors): boxes, balls, doors, and keys. These entities are placed in rooms of $8 \times 8 $ tiles that are connected by doors that can be locked or closed. The grid is procedurally generated (i.e. objects populating an episode are randomly chosen and their position, as well as the agent's position, are also random). Some of the objects are useful for the task to achieve, while others are considered as distractors (objects can't be crossed, the agent has to either bypass them or move them). The agent can do $6$ primitive actions: \texttt{turn left}, \texttt{turn right}, \texttt{go forward}, \texttt{toggle}, \texttt{pick up} to solve the language instruction (for instance \texttt{Pick up the red box}). To observe its environment, the agent has access to a partial view (i.e. it only sees the objects that belong to the $6 \times 6$ grid in front of it). BabyAI proposed to access this partial view through a symbolic mapping that returns $3$ matrices of size $6 \times 6$. The first matrix contains which object is in the observed cells, the second gives the color of these objects, and the last one their state (e.g. locked, open). When the agent completes the task after $N$ steps, it receives the reward $r_N = 1-0.9 \frac{N}{H}$, where $H$ is the maximum number of steps. During training, we multiply all rewards by $20$ to ensure a good propagation of the rewards as per \citep{mirchandani-etal-2021-ella}. If the agent has not completed the task in the current step, the reward is $0$. Additionally, BabyAI also provides visualization tools for experimenters to observe the grid and better grasp agents' behaviors. 

\subsection{BabyAI-Text}
\label{appendix:babyai_text}
BabyAI-Text is a textual environment that encapsulates BabyAI and provides a description of each observation instead of a symbolic representation. A textual description consists of a list of template descriptions with the following structure:

\par\smallskip $\bullet$ "You see a $<$\textit{object}$>$ $<$\textit{location}$>$" if the object is a key, a ball, a box or a wall.
\par\smallskip $\bullet$ "You see a(n) \textit{open/closed} door $<$\textit{location}$>$" , if the agent sees a door.
\par\smallskip $\bullet$ "You carry a $<$\textit{object}$>$", if the agent carries an object.

The $<$\textit{object}$>$, is composed of an \textit{adjective} (among $6$ possible colours: red, blue, green, yellow, grey, purple) and a noun (among $4$ possible: key, door, box, ball). The $<$\textit{location}$>$ is given as the number of steps \textit{right}, \textit{left}, and or \textit{forward} from the agent to the object. We illustrate this in the leftmost observation of \figurename~\ref{fig:babyai-text_schema} where the "yellow box" is "2 steps left and 1 step forward" from the agent (the red triangle). Thus an object described as "1 step forward" is right in front of the agent that does not need to \textit{go forward} if it wants to pick that object. Walls of the room are the only spatially extended objects in BabyAI-Text. We give their location at the closest distance to the agent. See the leftmost image of \figurename~\ref{fig:babyai-text_schema} for an example where the agent sees a wall "2 steps forward" and another wall "2 steps left". All of the choices for describing the environment constitute what we call the geometry of the environment, that the agent has to ground in order to succeed in the task. The presence of a fine grained geometry (with distances in steps to the different object in the room) is one of the main differences from other textual games such as TextWorld or ScienceWorld where all objects in a room are not spatially described.

Thanks to this extension, BabyAI-Text resembles a TextWorld (i.e. provides text descriptions of the observation and executes text commands) while keeping the inner minigrid environment along with BabyAI's tasks and visualization tools. Moreover, as our extension simply provides an alternative mapping of observations, one can both use and compare agents that either expect text-only observations (with BabyAI-Text) or symbolic observations (with BabyAI).

\newpage
\section{Additional results} \label{appendix:additional_results}

\subsection{Per-task success rate}
\label{appendix:success_per_task}

\begin{figure}[H]
    \vskip -0.2in
    \begin{center}
        \centerline{\includegraphics[width=0.6\textwidth]{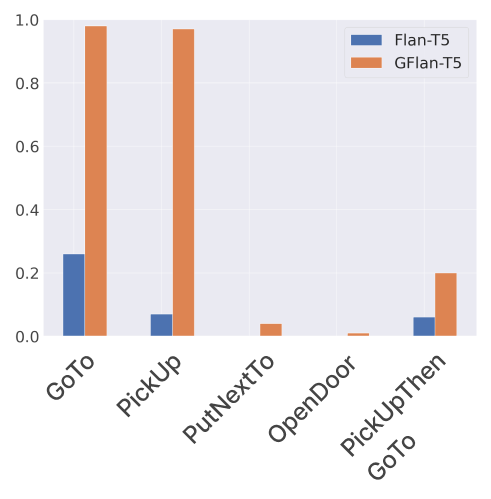}}
        \caption{Per-task success rate for the $1000$ evaluation trajectories performed in \figurename~\ref{fig:full_gene}.}
        \label{fig:per_task}
    \end{center}
    \vskip -0.2in
\end{figure}

In order to get a better understanding of our agent's capabilities, we report in \figurename~\ref{fig:per_task} the success rate on each task from our “no-change” evaluation in \figurename~\ref{fig:full_gene} (assessing the post-training performance of agents on $1000$ test episodes of our mixed setup) of Flan-T5 and GFlan-T5. These results (with $4$ seeds after $1.5$ million training steps) show that our functional grounding leads GFlan-T5 to master the \textit{GoTo} and \textit{PickUp} tasks while improving results on \textit{PutNextTo} and \textit{PickupThen/AfterPickup}. However, GFlan-T5 has not found yet any robust strategy for the  \textit{OpenDoor} task (being the hardest as the agent must find the right key and discover that the action “toggle” opens the door) in the relatively short allocated time.

\subsection{Averaging success rate over the tasks}
\label{appendix:average_sr_over_goals}

In our environments, goals are divided into five types of tasks of varying difficulty. Each goal is sampled randomly from the tasks when the environment is reset. We obtain the success rate for the goals at update \textit{u} by averaging over the completed trajectories during the collection phase. As several environments run in parallel, goals from easier tasks that are completed more quickly, such as \textit{GoTo} and \textit{PickUp}, tend to be more represented in the buffer.

\subsection{Textual vs symbolic representation} \label{appendix:additional_results_representation_impact}
In order to understand how the structure of the observation (i.e. either symbolic image using $3$ matrices containing integers defining respectively the object seen, its color and property if any or text) influences the success rate of an RL agent, we compare the DRRN and Symbolic-PPO respectively trained on BabyAI-Text and BabyAI on the \texttt{Go To Red Ball} task. In this task, the agent has to go in front of a red ball in $1$ room without any distractor (i.e. the task never changes, only the position of agent and red ball do). The task has been voluntarily chosen as trivial so that the main difference only comes from the way the information is given to the agent. Both the DRRN and Symbolic-PPO agents have a similar number of parameters (~$1$M), they both use recurrent layers to deal with partial observability and use the canonical action space. 

\begin{figure}[H]
    \vskip 0.2in
    \begin{center}
        \centerline{\includegraphics[width=0.6\textwidth]{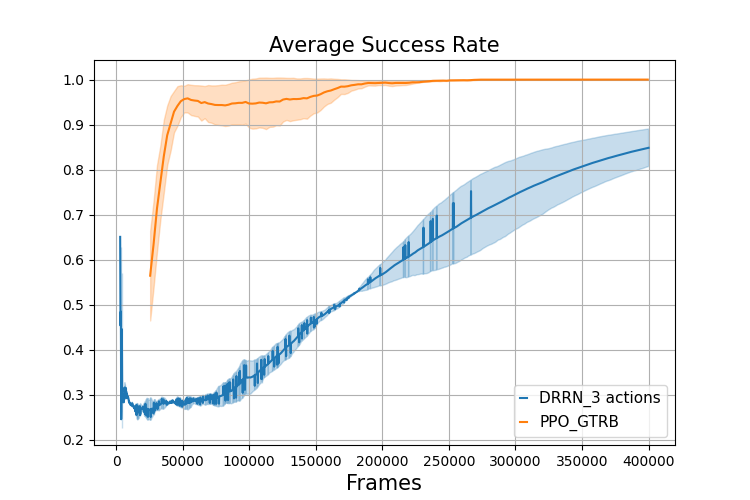}}
        \caption{Average success rate for DRRN and Symbolic-PPO on the \textit{Go To Red Ball} task with standard deviation over two random seeds. The PPO receives symbolic information and the DRRN gets textual observations.}
        \label{fig:compare_gtrb_drrn_ppo}
    \end{center}
    \vskip -0.2in
\end{figure}

Contrary to what one might assume in \figurename~\ref{fig:compare_gtrb_drrn_ppo} the PPO agent converges faster than the DRRN agent on this trivial task. Thus, symbolic observations make the learning easier for the agent. We conclude that even if language contains high-level information, understanding the link between spatial information and language is far more difficult than using symbolic information given in a matrix. Indeed, the matrix already contains a geometric bias favorable to the agent. We also want to point out that the DRRN is an off-policy RL method compared to PPO (which is on-policy) and that consequently, the DRRN was expected to be, by-design, more sample efficient.

\subsection{Impact of pretraining} \label{appendix:additional_results_impact_pretraining}
We test how pretraining structured our LLM allowing for efficient finetuning. We vary which weights of Flan-T5 are kept pretrained as well as how we compute actions' probability (i.e. either using our method reusing language modeling heads or using new action heads with an MLP). We evaluate the performance of $5$ models:
\par
\begin{figure}[H]
    \vskip 0.2in
    \begin{center}
        \centerline{\includegraphics[width=0.65\textwidth]{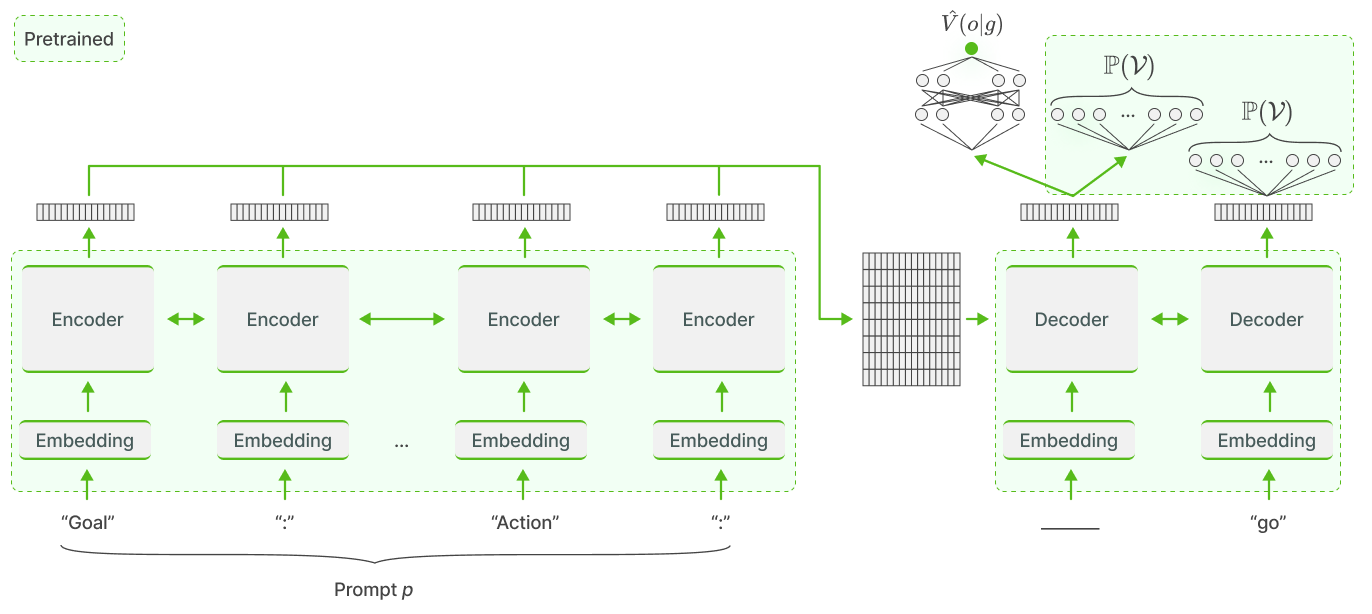}}
        \caption{GFlan-T5: We use the Flan-T5 architecture and add a value head. We initialize the agent with the pretrained weights (framed in green in the diagram) including its language modeling heads to compute action probabilities. The weights of the value head are initialized randomly. GFlan-T5 stands for: grounded Flan-T5.}
        \label{fig:GFlan-T5}
    \end{center}
    \vskip -0.2in
\end{figure}

\smallskip $\bullet$ The full LLM is pretrained and language modeling heads are used for actions probability: GFlan-T5 (\figurename~\ref{fig:GFlan-T5})
\par
\begin{figure}[H]
    \vskip 0.2in
    \begin{center}
        \centerline{\includegraphics[width=0.65\textwidth]{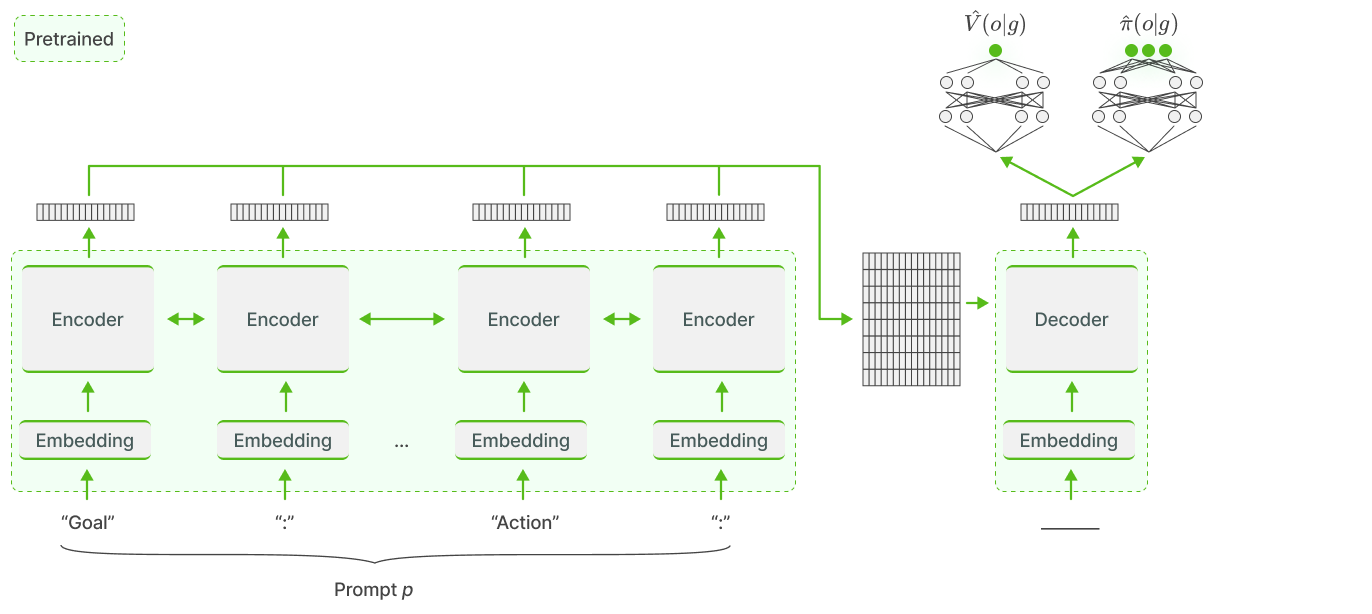}}
        \caption{AFlan-T5: We use the Flan-T5 architecture but replace the language modeling heads with action heads (that return the probability for each action) and add a value head. We initialize the embedding, the encoder and decoder parts of the agent with the pretrained weights (framed in green in the diagram) and the other weights randomly. AFlan-T5 stands for action heads Flan-T5.}
        \label{fig:AFlan-T5}
    \end{center}
    \vskip -0.2in
\end{figure}

\smallskip $\bullet$ The full LLM is pretrained and new action heads are used: AFlan-T5 (\figurename~\ref{fig:AFlan-T5})
\par
\begin{figure}[H]
    \vskip 0.2in
    \begin{center}
        \centerline{\includegraphics[width=0.65\textwidth]{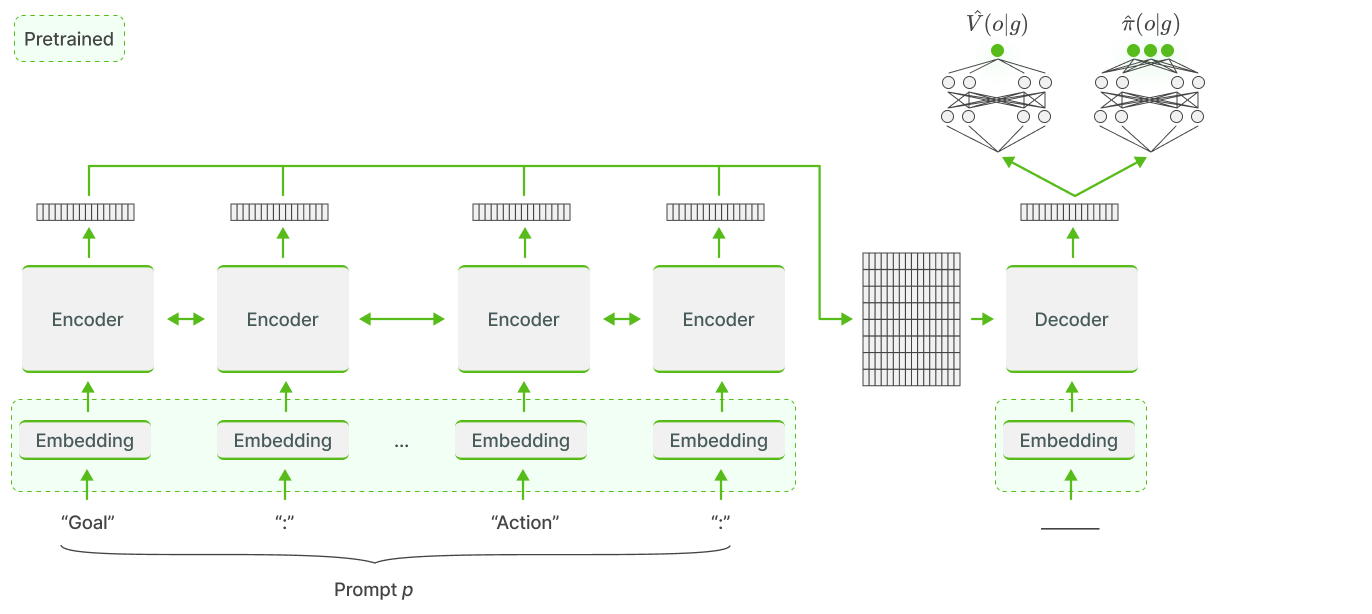}}
        \caption{NPAE-Flan-T5: We use the Flan-T5 architecture but replace the language modeling heads by action heads and add a value head. We initialize the embedding with the pretrained weights (framed in green in the diagram) and the other weights randomly. NPAE-Flan-T5 stands for: non-pretrained with action heads and pretrained embedding Flan-T5.} 
        \label{fig:NPAE-Flan-T5}
    \end{center}
    \vskip -0.2in
\end{figure}

\smallskip $\bullet$ Only the embedding layer's weights are kept pretrained (the rest of the LLM is randomly initialized) and new action heads are used: NPAE-Flan-T5 (\figurename~\ref{fig:NPAE-Flan-T5})
\par
\begin{figure}[H]
    \vskip 0.2in
    \begin{center}
        \centerline{\includegraphics[width=0.65\textwidth]{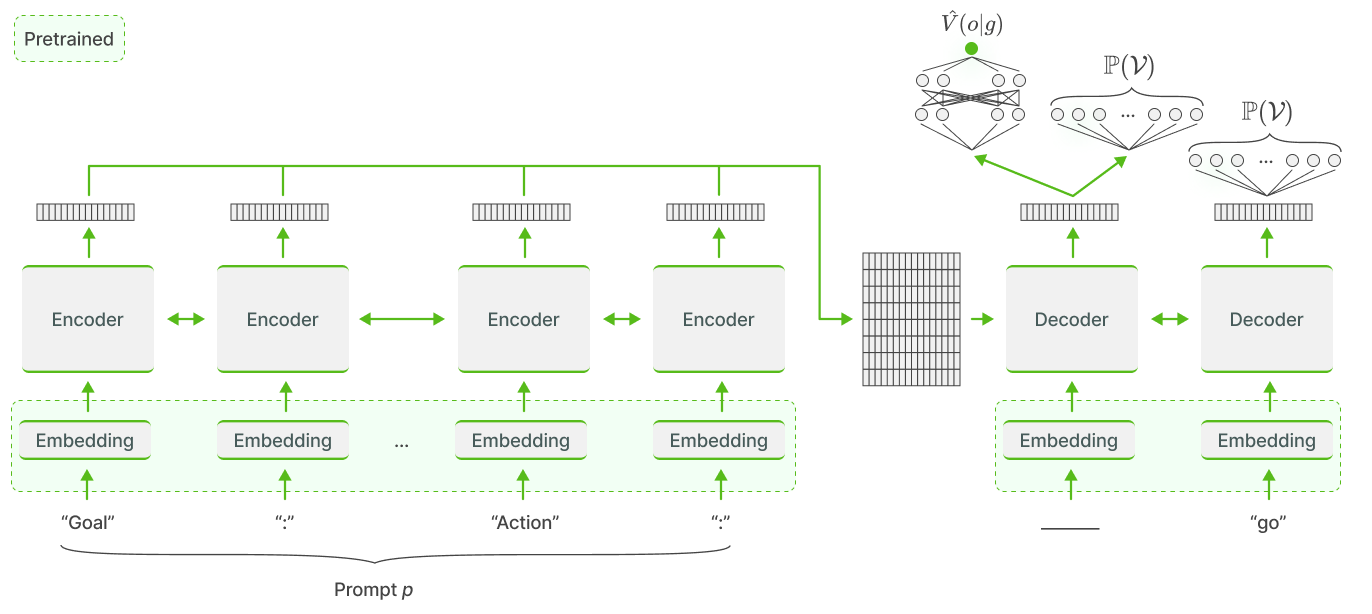}}
        \caption{NPE-Flan-T5: We use the Flan-T5 architecture and add a value head. We initialize the embedding with the pretrained weights (framed in green in the diagram) and the other weights randomly. NPE-Flan-T5 stands for: non-pretrained with pretrained embedding Flan-T5.}
        \label{fig:NPE-Flan-T5}
    \end{center}
    \vskip -0.2in
\end{figure}

\smallskip $\bullet$ Only the embedding layer's weights are kept pretrained (the rest of the LLM is randomly initialized) and the (randomly initialized) language modeling heads are used for actions' probability: \textbf{NPE-FlanT5} (\figurename~\ref{fig:NPE-Flan-T5})
 \par
\begin{figure}[H]
    \vskip 0.2in
    \begin{center}
        \centerline{\includegraphics[width=0.65\textwidth]{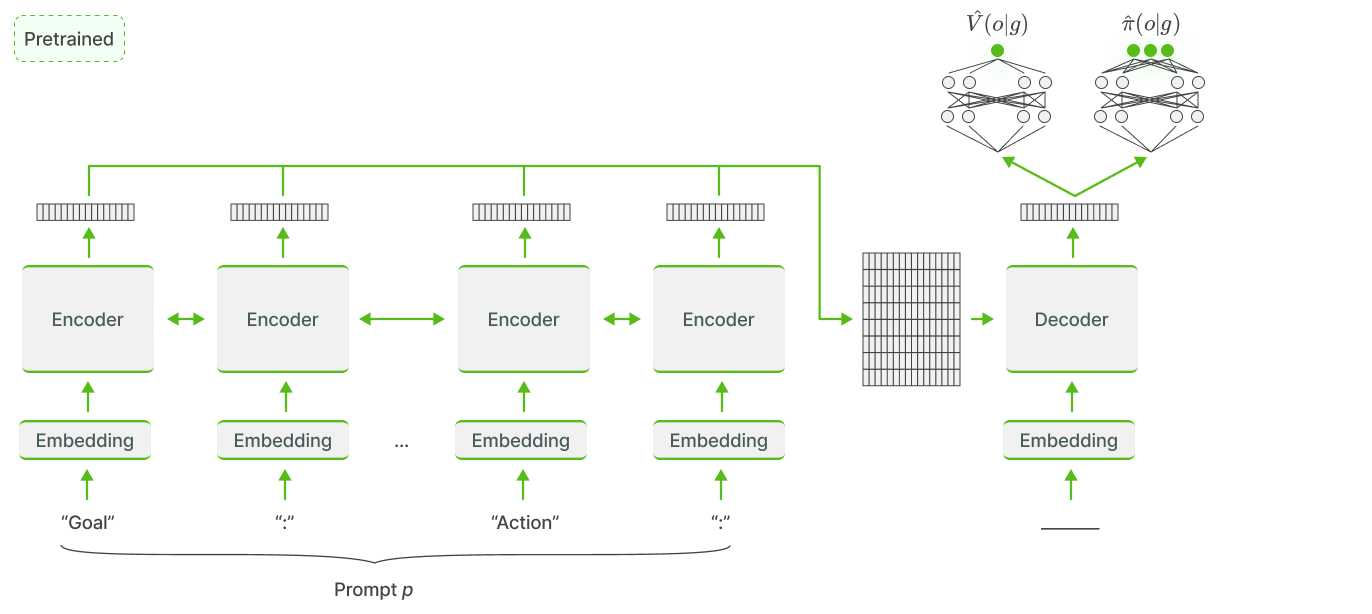}}
        \caption{NPA-Flan-T5: We use the Flan-T5 architecture but replace the language modeling heads with action heads and add a value head. We initialize all the weights randomly. NPA-Flan-T5 stands for: non-pretrained with action heads Flan-T5.}
        \label{fig:NPA-Flan-T5}
    \end{center}
    \vskip -0.2in
\end{figure}

\smallskip $\bullet$ All LLM's weights are randomly initialized and action heads are used: \textbf{NPA-Flan-T5} (\figurename~\ref{fig:NPA-Flan-T5})

\begin{figure}[H]
    \vskip 0.2in
    \begin{center}
        \centerline{\includegraphics[width=0.6\textwidth]{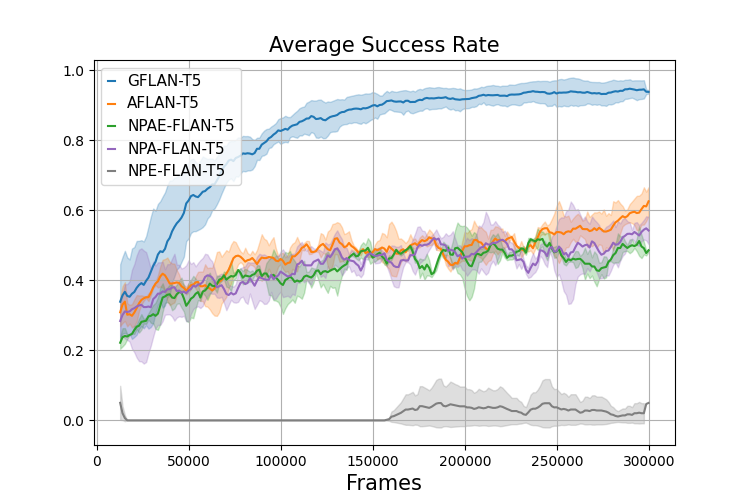}}
        \caption{Average success rate of varying pretrained weights and scoring method with standard deviation over two random seeds. We train all LLMs on the \textit{Go to $<$object$>$} task in $1$ room, with $8$ distractors, the $6$ canonical actions and using Flan-T5 large ($780$ million parameters) as architecture.}
        \label{fig:impact_pretraining_gtl}
    \end{center}
    \vskip -0.2in
\end{figure}

\figurename~\ref{fig:impact_pretraining_gtl} compares the training curves of the agents above on the task \textit{Go To $<$object$>$}. GFlan-T5 has unsurprisingly the best results as it is fully pretrained. More surprisingly, AFlan-T5 takes more steps than expected to perform better than the non-pretrained networks ($250000$ frames). We hypothesize that during the pertaining, the last transformer layer encodes information in a space designed for language modeling heads ($\approx 32000$ heads) which is not convenient for the non-pretrained $6$ actions heads. Indeed, AFlan-T5 has to make sense of this space before getting the benefits of having the rest of the network trained. This could explain why it suddenly performs better after $250000$ steps. Comparing NPAE, NPA and NPE Flan-T5, we see that the presence of an action head is crucial for non-pretrained networks. Indeed, the NPE fails to learn in the given number of steps compared to NPAE and NPA that have similar learning curves. A possible explanation is that for NPE, the information flow that is backpropagated through the gradient is really small due to the huge number of language modeling heads and the few number of tokens updated ($<100$). On the opposite, GFlan-T5, that also uses language modeling heads but is fully pretrained, only needs a light finetuning for the necessary tokens explaining its high success rate and sample efficiency.


\vspace{-1cm}
\subsection{Impact of the size of the LLM} \label{appendix:additional_results_llm_size}
The capacities of LLMs depend strongly on their size \citep{kaplan-etal-2020-scaling} and many properties of these networks only appear when they are large enough \citep{wei-etal-2022-emergent}. We consequently test the influence of the size of the LLM on our results by training $3$ different GFlan-T5 (as well as the DRRN and Symbolic-PPO baselines) on the \textit{Go to $<$object$>$} task for $400.000$ steps: GFlan-T5 \textbf{small} ($80$ million parameters), GFlan-T5 \textbf{large} ($780$ million parameters) and GFlan-T5 \textbf{XL} ($3$ billion parameters).

\begin{figure}[H]
    \vskip 0.2in
    \begin{center}
        \centerline{\includegraphics[width=0.6\columnwidth]{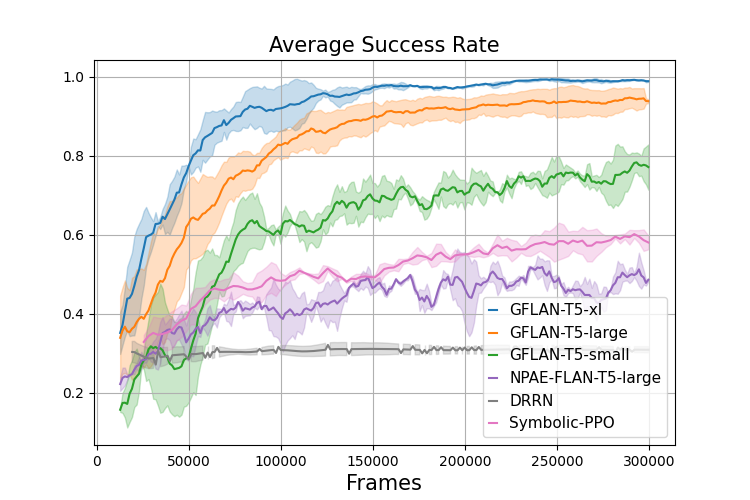}}
        \caption{Impact of the size of the LLM on online RL finetuning. We conduct the tests with the \textit{Go to $<$object$>$} task in $1$ room, with $8$ distractors. We measure the evolution of average success rate over $2$ seeds with standard deviation for GFlan-T5 \textbf{small} ($80$ million parameters), GFlan-T5 \textbf{large} ($780$ million parameters) and GFlan-T5 \textbf{XL} ($3$ billion parameters). DRRN, NPAE-Flan-T5-large and Symbolic-PPO are given as baselines.}
        \label{fig:impact_size_gtl}
    \end{center}
    \vskip -0.2in
\end{figure}

We show the evolution of average success rate over $2$ seeds in \figurename~\ref{fig:impact_size_gtl} highlighting that pretraining prior knowledge only looks impactful when the network is large enough. The difference between the learning properties of small and large models relates to the definition of an emergent behavior given by \citet{wei-etal-2022-emergent}: \textit{"an ability is emergent if it is not present in smaller models but is present in larger models"}. Beyond the data on which a model has been trained, the size of this model seems crucial for the acquisition of new knowledge about relations between entities during the finetuning phase.

\subsection{Impact of varying action space and distractors} \label{appendix:additional_results_action_space_distractors}
In this section, we detail the studies about the impact of varying the action space and the number of distractors.
\vskip -1cm
\subsubsection{Impact of the dimension of the action space}
\label{appendix:additional_results_action_space}
One of the expected advantages of pretrained LLMs in RL is that they avoid the \textit{Tabula-Rasa} paradigm and already have useful biases. In these experiments, we test the sensibility of LLMs to the size of the action space by using $3$ different action spaces (\textbf{restricted}, \textbf{canonical}, \textbf{augmented}) when trained on the \textit{Go to $<$object$>$} task.

\begin{figure}[H]
    \vskip 0.2in
    \begin{center}
        \centerline{\includegraphics[width=0.95\textwidth]{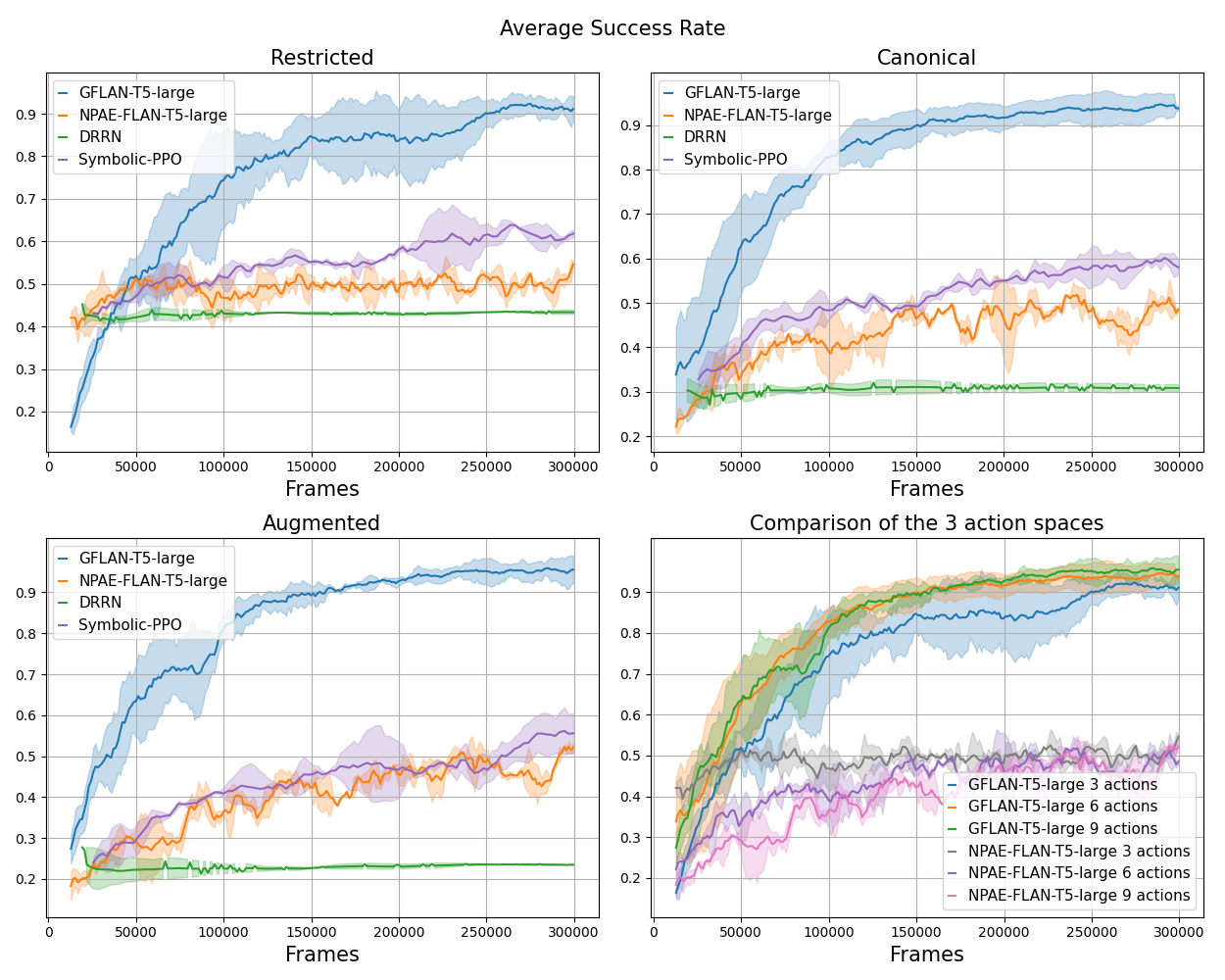}}
        \caption{Learning curves for the agents on the \textit{Go To} task for different sizes of action space (Restricted: $3$ actions, Canonical: $6$, Augmented: $9$, with only the $3$ actions that are useful). The success rate is given over $2$ seeds along with standard deviation.}
        \label{fig:ablation_sr_action_space}
    \end{center}
    \vskip -0.2in
\end{figure}

We conduct the tests in an environment with $1$ room, $8$ distractors and in \figurename~\ref{fig:ablation_sr_action_space} report full learning curves used to draw \figurename~\ref{fig:ablation_se_action_space}. We show that GFlan-T5 efficiently handles the different action spaces compared to the other agents. Its initial biases are particularly helpful when the action space is large. Indeed, when we look at the difference of sucess rate between GFlan-T5 and the second best-performing agent after the $50000$ first steps, there is almost no difference in the restricted settings and $0.35$ in the augmented settings. That supports the hypothesis that the results are due to LLMs' ability to discard useless action quickly at the beginning of finetuning. 

\subsubsection{Impact of the number of distractors}
\label{appendix:additional_results_distractors}

\begin{figure}[H]
    \vskip 0.2in
    \begin{center}
        \centerline{\includegraphics[width=0.95\textwidth]{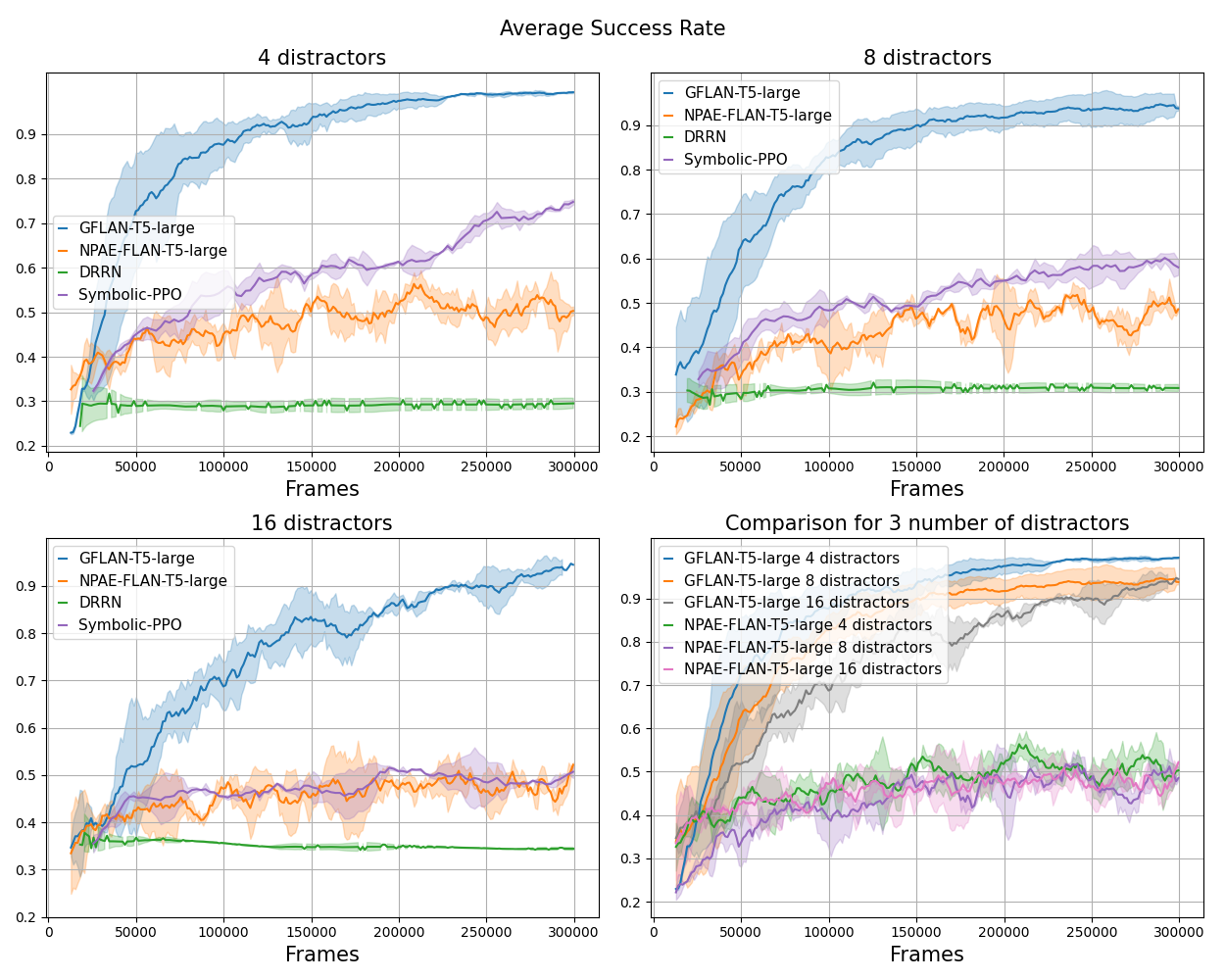}}
        \caption{Learning curves for the agents on the \textit{Go To} task for different number of distractors (4, 8, 16). The success rate is given over $2$ seeds with standard deviation.}
        \label{fig:ablation_sr_nbr_dist}
    \end{center}
    \vskip -0.2in
\end{figure}

In \figurename~\ref{fig:ablation_se_nbr_dist} we have shown that LLMs are less sensitive to variations on task complexity by plotting the evolution of sample efficiency (Equation~\eqref{eq:se}) for different numbers of distractors: $4$, $8$ and $16$. In \figurename~\ref{fig:ablation_sr_nbr_dist} we report the full learning curves. We observe that Symbolic-PPO's performances collapse when we go from $4$ to $16$ distractors whereas GFlan-T5's performances remain similar. NPAE-Flan-T5's performances are also non-affected by the change in the number of distractors but in this case we suppose it is because it cannot learn the task in $400000$ steps.

\subsection{Robustness to domain-specific vocabulary} \label{appendix:arobustness_domaine_specific_vocabulary}

\begin{figure}[H]
    \vskip 0.2in
    \begin{center}
        \centerline{\includegraphics[width=0.8\textwidth]{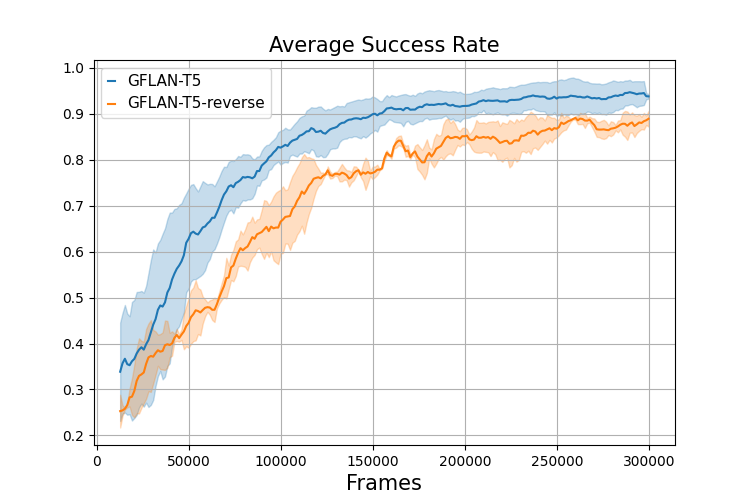}}
        \caption{Comparison of the average success rate over training for GFlan-T5 on the \textit{Go To} when the actions “turn left” and “turn right” are flipped ("-reverse"). The success rate is given over $2$ seeds with standard deviation.}
        \label{fig:inversed_left-right}
    \end{center}
    \vskip -0.2in
\end{figure}

In Section~\ref{sec:results_q1}, we have shown the robustness of our method to random vocabulary for words that do not influence the grounding of actions (in our case, the objects and their colors). Nonetheless, one can imagine an environment with a specific vocabulary where common words are used to describe particular technical terms with possibly very different meanings. To verify the impact of such environment on our training process, we trained GFlan-T5 on the \textit{GoTo} task where the actions “turn left” and “turn right” are flipped (i.e. using “turn left” makes the agent rotate to the right, and the opposite for “turn right”). \figurename~\ref{fig:inversed_left-right} shows that, while the prior knowledge of the LLM leads to poorer performance at the very beginning of training (as the LLM must learn to rotate left and right), GFlan-T5 converges at a similar speed than in the  non-flipped environment. This result hints robustness that our grounding method for LLMs can adapt to domain-specific vocabularies.

\newpage
\section{Evolution of actions distribution on evaluation prompts} 
\label{appendix:prompt_examples}
\begin{figure}[H]
    \vskip 0.2in
    \begin{center}
        \centerline{\includegraphics[width=\textwidth]{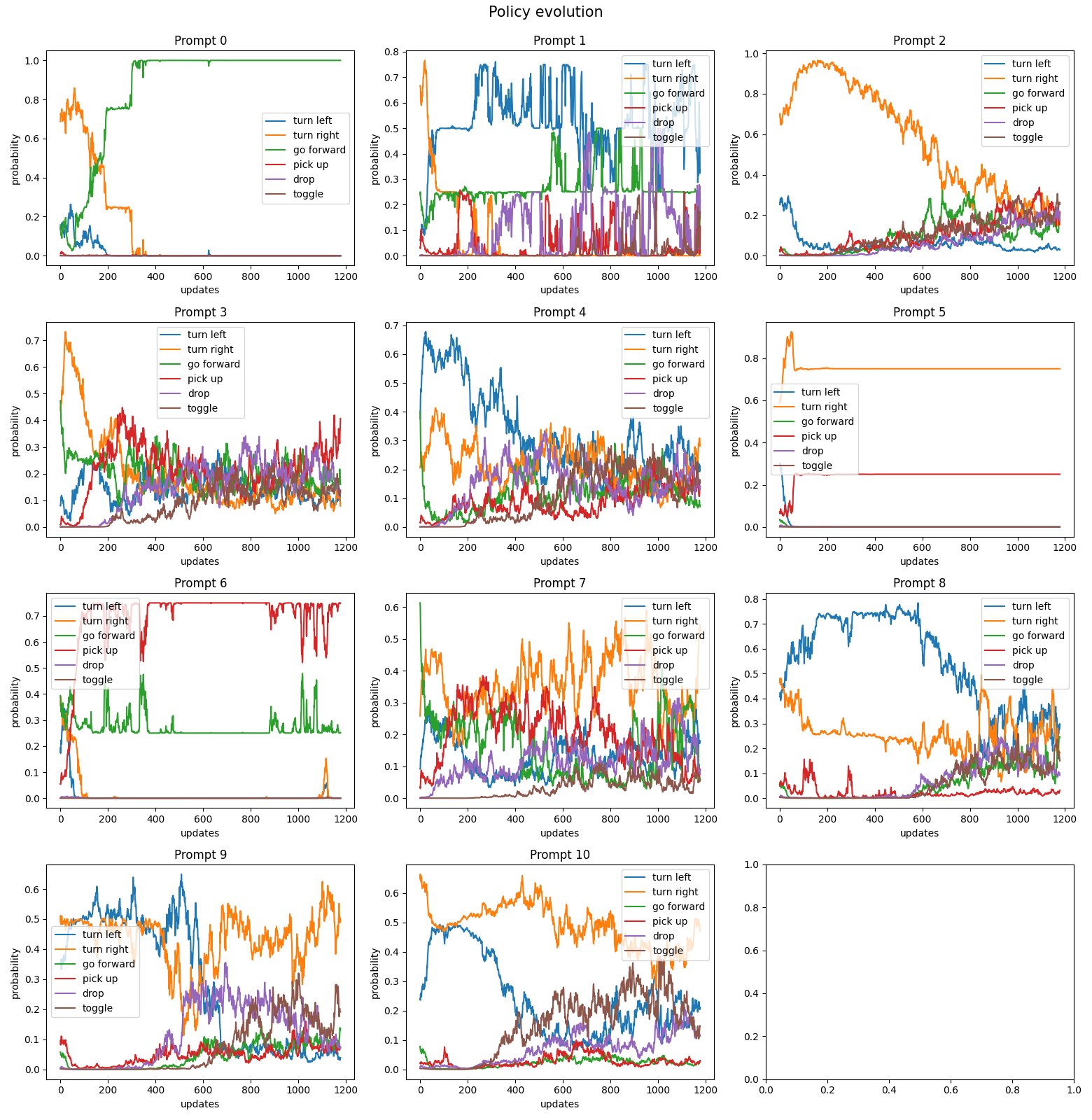}}
        \caption{Evolution of actions' probability over training for test prompts listed in Table \ref{tab:prompt}.}
        \label{fig:policy_evolution}
    \end{center}
    \vskip -0.2in
\end{figure}

To better grasp the skill acquisition dynamics when performing online RL grounding on GFlan-T5 in the multi-task setting of Section \ref{sec:results_q1}, we test at each update the LLM on $11$ prompts listed in Table \ref{tab:prompt}. We plot in \figurename~\ref{fig:policy_evolution} the evolution of action probabilities outputted by our LLM aiming to partially decipher the changes in the LLM and visualize which skill is acquired when.

Prompts \hyperlink{prompt_0}{0} and \hyperlink{prompt_1}{1} are simple navigation tasks. The agent has to move in the direction given in the prompt. Looking at the corresponding plots we observe two things: first, the optimal behavior is learned in less than a hundred of updates, even for prompt \hyperlink{prompt_1}{1} for which the bias at the beginning is both wrong and high. Second, from the beginning, only the navigation actions (\texttt{turn left}, \texttt{turn right}, \texttt{go forward}) relevant for the \textit{Go To $<$object$>$} task have a high probability. Therefore, the Flan-T5 $780$M seems to already have useful biases for navigation and is able to quickly update or correct them through interactions.

We observe similar useful biases with the \textit{Pick Up $<$object$>$} task (using prompts \hyperlink{prompt_5}{5} and \hyperlink{prompt_6}{6}). Indeed, at the beginning, both the \texttt{pick up} action and navigation actions already have a high probability. 

We can see that the agent struggles to ground the geometry of the environment with prompts \hyperlink{prompt_6}{6} and \hyperlink{prompt_7}{7}. Indeed, it has to understand that an object described at "1 step forward" is in front of it such that it can pick it up or drop it directly without moving further. While GFlan-T5 eventually seems to understand it, it still shows some hesitation as proven by the fact that it gives almost the same probability to \texttt{go forward} and \texttt{pick up} or \texttt{drop} for the prompt \hyperlink{prompt_7}{7} at the end of training (see \figurename~\ref{fig:policy_evolution}).

We also verify how GFlan-T5 understands temporal constructions such as doing an action A then an action B (prompt \hyperlink{prompt_8}{8}) or doing an action A after doing an action B (prompt \hyperlink{prompt_9}{9}). These two test prompts are exactly the same except for the goal where prompt \hyperlink{prompt_8}{8} uses "then" and prompt \hyperlink{prompt_9}{9} uses "after" to link the two actions. We observe that when the order of actions in the task specification is the same as the one the agent has to do (i.e. prompt \hyperlink{prompt_8}{8}), the LLM quickly and learns to choose the right action even if during the learning it loses its ability (with \texttt{turn left} and \texttt{turn right} that are almost at the same probability. However, when the order of actions mentioned in the goal specification is reversed (prompt \hyperlink{prompt_9}{9}), the LLM ends up favoring the wrong direction and exhibits much more hesitation from the begining of the training. This qualitative observation concurs with the measure of success rate given in Appendix~\ref{appendix:tab_groud_after_then}.

The prompts \hyperlink{prompt_2}{2}, \hyperlink{prompt_3}{3} and \hyperlink{prompt_4}{4} show that the agent has difficulties with the task \textit{Open $<$door$>$}. This task is fairly complex since the agent has to infer that a key of the same color as the closed door is required to open it. In the given training budget, the agent fails to associate the need of a key with the task. 

Finally we test the agent on a task that is not seen during training. It is the generalization task \textit{Pick up $<$object A$>$ then/after Pick up $<$object B$>$} from Q3 Section~\ref{sec:results_q3}, composed from two tasks seen during training \textit{Pick up} and \textit{Pick up then Go To} (prompt \hyperlink{prompt_10}{10}). The prompt is built such that the agent has accomplished half of the instruction and has to drop the object it carries in order to pick another one. The action \texttt{drop} is the optimal one because it is the only one that allows the agent to complete the goal in a minimum number of steps. Between the updates $400$ and $600$ the agent begins to increase the probability of the \texttt{drop} action. This change is correlated to the change of distribution in prompt \hyperlink{prompt_7}{7}. It can be interpreted as the fact that the action \texttt{drop} begins to be grounded after $800$ updates.

\newpage
\setlength{\LTleft}{-20cm plus -1fill}
\setlength{\LTright}{\LTleft}
\begin{longtable}{|l|l|l|l|}
    \caption{Test prompts. The prompts' header (\textit{Possible action of the agent: turn left, turn right, go forward, pick up, drop, toggle}) is not shown below as it remains the same for all prompts} \label{tab:prompt} \\
    
    \toprule
    \textbf{Ids} & \textbf{Tasks} & \textbf{Prompts} \hypertarget{prompt_0}  & \textbf{Comments} \\
    \midrule
    \endfirsthead

    \multicolumn{4}{c}%
    {{\bfseries \tablename\ \thetable{} -- continued from previous page}} \\
    \toprule
    \textbf{Id} & \textbf{Task} & \textbf{Prompt} & \textbf{Comments} \\
    \midrule 
    \endhead

    \hline \multicolumn{4}{|r|}{{Continued on next page}} \\ \hline
    \endfoot
    
    \bottomrule
    \endlastfoot
    
    \multirow{10}{*}{$0$} & \multirow{10}{*}{Go To $<$object$>$} & {\textbf{Goal of the agent}}: go to the green ball & Simple navigation task. \\
    & & \textbf{Observation 0}: You see a wall 2 step left, You see a purple key 1 step left & \\
    & & and 2 steps forward, You see a yellow key 1 step left and 1 step forward, & \\
    & & You see a green ball 3 steps forward, You see a grey ball 1 step right and & \\
    & & 5 steps forward, You see a green key 1 step right and 2 steps forward, & \\
    & & You see a grey ball 1 step right and 1 step forward, You see a green key & \\
    & & 2 steps right and 4 steps forward, You see a red box 2 steps right and & \\
    & & 2 steps forward, & \\
    & & \textbf{Action 0}: & \\
    & & \textbf{Expected answer: go forward} \hypertarget{prompt_1} & \\
    \hline
    
    \multirow{22}{*}{$1$} & \multirow{22}{*}{Go To $<$object$>$} & \textbf{Goal of the agent}: go to the green ball & Simple navigation task. \\
    & & \textbf{Observation 0}: You see a wall 2 step left, You see a purple key 1 step left & \\
    & & and 2 steps forward, You see a yellow key 1 step left and 1 step forward, & \\
    & & You see a green ball 3 steps forward, You see a grey ball 1 step right and & \\
    & & 5 steps forward, You see a green key 1 step right and 2 steps forward, & \\
    & & You see a grey ball 1 step right and 1 step forward, You see a green key & \\
    & & 2 steps right and 4 steps forward, You see a red box 2 steps right and & \\
    & & 2 steps forward, & \\
    & & \textbf{Action 0}: go forward & \\
    & & \textbf{Observation 1}: You see a purple key 1 step left and 1 step forward, & \\ 
    & & You see a yellow key 1 step left, You see a green ball 2 steps forward, & \\
    & & You see a grey ball 1 step right and 4 steps forward, You see a green key & \\
    & & 1 step right and 1 step forward, You see a grey ball 1 step right, You see & \\
    & & a green key 2 steps right and 3 steps forward, You see a red box 2 steps right & \\
    & & and 1 step forward, & \\
    & & \textbf{Action 1}: turn right & \\
    & & \textbf{Observation 2}: You see a wall 2 step right, You see a green key 3 steps left & \\ 
    & & and 2 steps forward, You see a green ball 2 steps left, You see a red box 1 step & \\
    & & left and 2 steps forward, You see a green key 1 step left and 1 step forward, & \\
    & & You see a grey ball 1 step forward, & \\
    & & \textbf{Action 2}: & \\
    & & \textbf{Expected answer: turn left} \hypertarget{prompt_2} & \\
    \hline
    
    \multirow{8}{*}{$2$} & \multirow{8}{*}{Open $<$adj$>$ door} & \textbf{Goal of the agent}: open the purple door & Inference task \\
    & & \textbf{Observation 0}: You see a wall 3 steps forward, You see a wall 3 steps left, & The agent has to infer \\
    & & You see a yellow key 1 step right and 1 step forward, You see a locked & that a key of the same \\
    & & purple door 2 steps right and 3 steps forward, You see a purple ball 3 steps & color is needed and \\
    & & right and 1 step forward, You see a green box 3 steps right, You see a purple key & moves toward it.\\
    & & 2 steps left, & \\
    & & \textbf{Action 0}:  & \\ 
    & & \textbf{Expected answer: turn left} \hypertarget{prompt_3} & \\
    \hline
    
    \multirow{14}{*}{$3$} & \multirow{14}{*}{Open $<$adj$>$ door} & \textbf{Goal of the agent}: open the purple door & Inference task \\
    & & \textbf{Observation 0}: You see a wall 3 steps forward, You see a wall 3 steps left, & The agent has to infer \\
    & & You see a yellow key 1 step right and 1 step forward, You see a locked & that a key of the same \\
    & & purple door 2 steps right and 3 steps forward, You see a purple ball 3 steps & color is needed and \\
    & & right and 1 step forward, You see a green box 3 steps right, You see a purple key & pick it up.\\
    & & 2 steps left, & \\
    & & \textbf{Action 0}: turn left & \\
    & & \textbf{Observation 1}: You see a wall 3 steps forward, You see a wall 3 steps right, & \\
    & & You see a purple key 2 steps forward, & \\
    & & \textbf{Action 1}: go forward & \\
    & & \textbf{Observation 2}: You see a wall 2 steps forward, You see a wall 3 steps right, & \\
    & & You see a purple key 1 step forward, & \\
    & & \textbf{Action 2}: & \\ 
    & & \textbf{Expected answer: pick up} \hypertarget{prompt_4} & \\
    \hline
    
    \multirow{14}{*}{$4$} & \multirow{14}{*}{Open $<$adj$>$ door} & \textbf{Goal of the agent}: open the purple door & Inference task \\
    & & \textbf{Observation 0}: You carry a purple key, You see a wall 3 steps forward, & The agent has to infer \\
    & & You see a wall 5 steps left, You see a yellow key 1 step left and 1 step forward, & that you can open a \\
    & & You see a locked purple door 3 steps forward, You see a purple ball 1 step right & closed door by toggling\\
    & & and 1 step forward, You see a green box 1 step right, & it while having a key \\
    & & \textbf{Action 0}: go forward & of the same color. \\
    & & \textbf{Observation 1}: You carry a purple key, You see a wall 2 steps forward, & \\
    & & You see a wall 5 steps left, You see a yellow key 1 step left, You see a & \\ 
    & & locked purple door 2 steps forward, You see a purple ball 1 step right, & \\
    & & \textbf{Action 1}: go forward & \\
    & & \textbf{Observation 2}: You carry a purple key, You see a wall 1 step forward, & \\
    & & You see a wall 5 steps left, You see a locked purple door 1 step forward, & \\
    & & \textbf{Action 2}: & \\ 
    & & \textbf{Expected answer: toggle} \hypertarget{prompt_5} & \\
    \hline
    
    \multirow{6}{*}{$5$} & \multirow{6}{*}{Pick up $<$object$>$} & \textbf{Goal of the agent}: pick up green box  & The agent has to \\
    & & \textbf{Observation 0}: You see a wall 2 steps forward, You see a wall 2 steps left, & reuse knowledge \\
    & & You see a yellow ball 1 step left and 1 step forward, You see a green box & from navigation task. \\
    & & 2 steps right, & \\
    & & \textbf{Action 0}:  & \\ 
    & & \textbf{Expected answer: turn right} \hypertarget{prompt_6} & \\
    \hline
    
    \multirow{13}{*}{$6$} & \multirow{13}{*}{Pick up $<$object$>$} & \textbf{Goal of the agent}: pick up green box  & The agent has to \\
    & & \textbf{Observation 0}: You see a wall 2 steps forward, You see a wall 2 steps left, & reuse knowledge \\
    & & You see a yellow ball 1 step left and 1 step forward, You see a green box & from navigation task. \\
    & & 2 steps right, & and understand the\\
    & & \textbf{Action 0}: turn right & geometry of the room. \\ 
    & & \textbf{Observation 1}: You see a wall 2 steps left, You see a blue key 1 step right, & \\
    & & You see a red ball 2 steps right and 1 step forward, You see a green box & \\
    & & 2 steps forward, & \\
    & & \textbf{Action 1}: go forward & \\
    & & \textbf{Observation 2}: You see a wall 2 steps left, You see a red ball 2 steps right, & \\
    & & You see a green box 1 step forward, & \\
    & & \textbf{Action 2}: & \\
    & & \textbf{Expected answer: pick up} \hypertarget{prompt_7} & \\
    \hline

    & & \textbf{Goal of the agent}: put blue ball next to red box  & The agent has to \\
    & & \textbf{Observation 0}: You carry a blue ball, You see a wall 5 steps forward, & reuse knowledge \\
    & & You see a wall 2 steps left, You see a grey key 1 step right and 2 steps & from navigation \\
    & & forward, You see a red box 3 steps forward, & and understand the \\
    $7$ & Put $<$object A$>$ & \textbf{Action 0}: go forward & geometry of the room. \\ 
    & next to $<$object B$>$ & \textbf{Observation 1}: You carry a blue ball, You see a wall 4 steps forward, & \\
    & & You see a wall 2 steps left, You see a grey key 1 step right and 1 step forward, & \\
    & & You see a red box 2 steps forward, & \\
    & & \textbf{Action 1}: & \\
    & & \textbf{Expected answer: drop} \hypertarget{prompt_8} & \\
    \hline

    & & \textbf{Goal of the agent}: pick up the blue ball then go to the red box  & Prompt 8 and 9 \\
    $8$ & Pick up $<$object A$>$ & \textbf{Observation 0}: You see a wall 3 steps forward, You see a wall 4 steps right, & test the ability \\
    & then go to $<$object B$>$ & You see a purple key 2 steps forward, You see a red box 2 steps right, & of the agent to \\
    & & You see a blue ball 2 steps left, & understand \\
    & & \textbf{Action 0}: & temporal concepts.\\ 
    & & \textbf{Expected answer: turn left} \hypertarget{prompt_9} & \\
    \hline

    & & \textbf{Goal of the agent}: go to the red box after you pick up the blue ball  & Same as prompt 8 \\
    & & \textbf{Observation 0}: You see a wall 3 steps forward, You see a wall 4 steps right, & but sentence action\\
    $9$ & Go to $<$object B$>$ & You see a purple key 2 steps forward, You see a red box 2 steps right, &  order different from\\
    & after you pick up & You see a blue ball 2 steps left, & execution order.\\
    & $<$object A$>$ & \textbf{Action 0}: & \\ 
    & & \textbf{Expected answer: turn left} \hypertarget{prompt_10}& \\
    \hline

    & & \textbf{Goal of the agent}: pick up the green key then pick up the red box  & Task never seen in \\
    & & \textbf{Observation 0}: You carry a green key, You see a wall 4 steps forward, & training to \\
    $10$ & Pick up $<$object A$>$ & You see a wall 4 steps left, You see a red box 1 step left, You see a purple ball & analyze generalization.\\
    & then pick up $<$object B$>$ & 2 steps left and 1 step forward, & \\
    & & \textbf{Action 0}: & \\ 
    & & \textbf{Expected answer: drop} & \\
\end{longtable}

\newpage
\section{Generalization tests details} \label{appendix:generalization_test}

\subsection{Recapitulating results table}
\label{appendix:recap_table}

In this section we summarize the numerical results shown in \figurename~\ref{fig:full_gene} with the confidence intervals calculated as explained in Appendix~\ref{appendix:confidence_interval}.
\begin{table}[htpb]
    \caption{Generalization tests}
    \label{tab:full_recap_tab_gene}
    \centering
    \begin{tabular}{l l|l l l l l}
        \toprule
        \multicolumn{2}{c}{\textbf{Environments}} &  \textbf{GFlan-T5} & \textbf{Flan-T5} & \textbf{NPAE} & \textbf{DRRN} & \textbf{Random} \\
        \midrule
        \multirow{5}{*}{\textbf{Q2}} & Mix - no change & $0.89 \pm 0.05$ & $0.11 \pm 0.03$ & $0.17 \pm 0.04$ & $0.14 \pm 0.02$ & \multirow{5}{*}{$0.15 \pm 0.05$} \\
        & Mix - out-of-vocabulary & \multirow{2}{*}{$0.87 \pm 0.05$} & \multirow{2}{*}{$0.09 \pm 0.02$} & \multirow{2}{*}{$0.16 \pm 0.05$} & \multirow{2}{*}{$0.15 \pm 0.00$} & \\
        & nouns & & & & & \\
        & Mix - invented nouns &  \multirow{2}{*}{$0.88 \pm 0.06$} &  \multirow{2}{*}{$0.11 \pm 0.03$} &  \multirow{2}{*}{$0.16 \pm 0.03$} &  \multirow{2}{*}{$0.16 \pm 0.00$} & \\
        & and adjectives & & & & & \\
        \midrule
        \multirow{4}{*}{\textbf{Q3}} & Pick up then/after & \multirow{2}{*}{$0.12 \pm 0.06$} & \multirow{2}{*}{$0.02 \pm 0.00$} & \multirow{2}{*}{$0.06 \pm 0.01$} & \multirow{2}{*}{$0.06 \pm 0.03$} & \multirow{2}{*}{$0.05 \pm 0.05$} \\
        &  pick up & & & & & \\
        & Mix - synonym actions & $0.12 \pm 0.12$ & $0.02 \pm 0.00$ & $0.16 \pm 0.04$ & $0.17 \pm 0.04$ & $0.15 \pm 0.05$ \\
         & Go To - English & $0.99 \pm 0.01$ & $0.27 \pm 0.03$ & $0.31 \pm 0.04$ & $0.31 \pm 0.03$ & \multirow{2}{*}{$0.30 \pm 0.05$} \\
         & Go To - French & $0.02 \pm 0.01$ & $0.03 \pm 0.00$ & $0.30 \pm 0.02$ & $0.31 \pm 0.02$ & \\
        \bottomrule
    \end{tabular}
\end{table}

\subsection{Complementary tests for Q2}
\label{appendix:generalization_test_q2}
\begin{table}[htpb]
    \caption{Complementary tests for Q2}
    \label{tab:complement_tests_Q2}
    \centering
    \begin{tabular}{l|l|l|l|l|l}
        \toprule
        \textbf{Environments} &  \textbf{GFlan-T5} & \textbf{Flan-T5} & \textbf{NPAE} & \textbf{DRRN} & \textbf{Random} \\
        \midrule
        Mix - no change & $0.89 \pm 0.05$ & $0.11 \pm 0.03$ & $0.17 \pm 0.02$ & $0.14 \pm 0.02$ & \multirow{5}{*}{$0.15 \pm 0.05$} \\
        Mix - unseen & \multirow{2}{*}{$0.87 \pm 0.03$}  & \multirow{2}{*}{$0.12 \pm 0.03$} & \multirow{2}{*}{$0.16 \pm 0.08$} & \multirow{2}{*}{$0.17 \pm 0.09$} & \\
        in-vocabulary objects & & & & & \\
        Mix - out-of-vocabulary  & \multirow{2}{*}{$0.87 \pm 0.07$} & \multirow{2}{*}{$0.16 \pm 0.03$} & \multirow{2}{*}{$0.16 \pm 0.02$} & \multirow{2}{*}{$0.16 \pm 0.01$} & \\
        adjectives & & & & & \\
        \bottomrule
    \end{tabular}
\end{table}

To further analyze results from in Section~\ref{sec:results_q2}, we conduct more systematic tests on different aspects of the generalization to new words. The results are given in Table~\ref{tab:complement_tests_Q2}. 

\paragraph{Unseen in-vocabulary objects}
During training we remove tasks whose goal contain the following objects: yellow box, red key, red door, green ball and, grey door. Nonetheless, the agent can have these objects as distractors and so have seen them during training. We assess how our agents perform on the mix of tasks with goals using only these objects. The success rate of $0.87$ points out that GFlab-T5 is unaffected by the use of unseen in-vocabulary objects.

\paragraph{Unseen out-of-vocabulary adjectives}
We perform the same test as for out-of-vocabulary nouns in Section~\ref{sec:results_q2} but this time with adjectives that do not belong to the BabyAI-Text vocabulary. We generate the prompt by exchanging the adjectives with predefined synonyms (see Table~\ref{tab:n_adj_mod}). Similarly to the test with out-of-vocabulary nouns, the test with out-of-vocabulary adjectives reveals that GFlan-T5 is unaffected by this change. Indeed, the success rate is of $0.87$ compared to the one of mix of tasks without change at $0.89$.


\subsection{Complementary tests for Q3}
\label{appendix:generalization_test_q3}
\begin{table}[htpb]
    \caption{Complementary tests for Q3}
    \label{tab:complement_tests_Q3}
    \centering
    \begin{tabular}{l|l|l|l|l|l}
        \toprule
        \textbf{Environments} &  \textbf{GFlan-T5} & \textbf{Flan-T5} & \textbf{NPAE} & \textbf{DRRN} & \textbf{Random} \\
        \midrule
        Go To - English & $0.99 \pm 0.01$ & $0.27 \pm 0.03$ & $0.31 \pm 0.04$ & $0.31 \pm 0.03$ & \multirow{4}{*}{$0.30 \pm 0.05$} \\
        Go To - French & $0.02 \pm 0.01$ & $0.03 \pm 0.00$ & $0.30 \pm 0.02$ & $0.31 \pm 0.02$ & \\
        Go To - English with & \multirow{2}{*}{$0.15 \pm 0.04$} & \multirow{2}{*}{$0.26 \pm 0.02$} & \multirow{2}{*}{$0.31 \pm 0.01$} & \multirow{2}{*}{$0.33 \pm 0.00$} & \\
        actions in French & & & & & \\
        \bottomrule
    \end{tabular}
\end{table}

In Section~\ref{sec:results_q3}, we observe that GFlan-T5 fails to generalize to an environment where we change the language. We hypothesize that such a change modifies too many grounded symbols at once. To verify this hypothesis, we test a middle-ground version, where we keep the environment in English but actions are in French. In this setting, Table~\ref{tab:complement_tests_Q3} shows that the success rate of the agent ($0.15$) is better than the fully french environment ($0.02$). This observation supports that finetuned agents tend to generalize to related words in other languages. Nonetheless, this ability seems highly dependent on the number of grounded words we modify.

\subsection{LLM grounding of temporal symbols: "then" and "after"}
\label{appendix:tab_groud_after_then}
In this experiment we observe the dynamics of functional grounding of instructions containing the temporal symbols "then" and "after" using the tasks: \textit{Pick up $<$object A$>$ then go to $<$object B$>$} and \textit{Go to $<$object B$>$ after pick up $<$object A$>$}. As the order of the action matters to have the task considered completed, a correct grounding of these symbols is crucial. Table~\ref{tab:then_after} shows that GFlan-T5 has a better grounding of these words than the original Flan-T5 agent. Moreover, we observe a slight bias after finetuning: the agent has stronger performances for the tasks with "then" (success rate of $0.22$) compared to the tasks with "after" (success rate of $0.17$). We hypothesize it is easier to ground the word "then" because the order of the actions the agent must do is the same as the order in which the actions appear in the instructions. A qualitative example of this behavior is given in Appendix~\ref{appendix:prompt_examples} (prompts \hyperlink{prompt_8}{8}, \hyperlink{prompt_9}{9}).

\begin{table}[htpb]
    \caption{Test on tasks with temporal components}
    \label{tab:then_after}
    \centering
    \begin{tabular}{l|l|l|l|l|l}
        \toprule
        \textbf{Environments} &  \textbf{GFlan-T5} & \textbf{Flan-T5} & \textbf{NPAE} & \textbf{DRRN} & \textbf{Random} \\
        \midrule
        Mix of tasks then/after & $0.23 \pm 0.06$ & $0.12 \pm 0.01$ & $0.09 \pm 0.01$ & $0.09 \pm 0.02$  & \multirow{3}{*}{$0.04 \pm 0.05$} \\
        Tasks with then only & $0.22 \pm 0.11$ & $0.12 \pm 0.01$ & $0.10 \pm 0.003$ & $0.10 \pm 0.02$ & \\
        Tasks with after only & $0.17 \pm 0.05$ & $0.13 \pm 0.05$ & $0.10 \pm 0.03$ & $0.10 \pm 0.01$ & \\
        \bottomrule
    \end{tabular}
\end{table}

\newpage
\section{Distributed experimental setup} \label{appendix:distributed}
In order to accelerate our online RL finetuning, we first leverage a classic distributed data collection setup where $32$ BabyAI-Text environments are running in parallel (all on CPUs). Our environments are run in a synchronous way, meaning that at every step, we get $32$ current states and need to send $32$ actions back to the environment. In very classic RL setups, policy networks are usually small and we simply batch the $32$ states, feed them to the network and obtain the $32$ actions' probability before sampling from them and choosing one action per environment. However, as explained in Section~\ref{sec:methods_interact}, our method requires $|\mathcal{A}|$ forward passes on a potentially very large and computationally expensive LLM in order to compute actions' probability for a single environment. Hence, we now need $32 \times |\mathcal{A}|$ forward passes for a single step in all environments, which can easily become a huge bottleneck in our training process.

To overcome this, we deploy for each of our experiments in Section \ref{sec:experiments} $4$ instances of our LLM all running in parallel. We load and use LLMs through the Hugging Face Transformers Python library\footnote{\url{https://huggingface.co/docs/transformers/index}}. Our method relies on a simple client-server architecture where the RL script acts as a client sending requests to LLMs. This client communicates with a master server which dispatches the call over multiple servers (i.e. one per LLM). Once each LLM has computed its subset of the call, the master gathers results and sends the response to the RL client. We use Pytorch Distributed\footnote{\url{https://pytorch.org/docs/stable/distributed.html}} with the GLOO backend for communication (hence possible both on CPU-only and GPU setups). We wrap all these in a Python library called \textit{Lamorel} which can dispatch calls over the deployed LLMs from a single line of code in the RL loop asking for actions' probability for all environments. Using this method, we observe a quasi-linear scaling with the number of deployed LLMs. 

Once transitions have been collected, we update our LLM using the PPO loss. For this, \textit{Lamorel} helps parallelize the gradients' computation with a Distributed Data Parallelism\footnote{\url{https://pytorch.org/tutorials/intermediate/ddp_tutorial.html}} setup where forward and backward passes over transitions are also dispatched on the different instances of our LLMs. Then, \textit{Lamorel} helps gather gradients and update each LLM (as well as their value head) the same way. In addition, \textit{Lamorel} also helps define a custom computational graph linked to the LLM. We use this to add MLPs on top of our Flan-T5 model for the value head (see experiments with action heads in Section \ref{appendix:additional_results_impact_pretraining}).

When using \textbf{Flan-T5 780M}, each LLM instance is distributed (Vertical Model Parallelism\footnote{Layers are spread across GPUs (\url{https://huggingface.co/docs/transformers/v4.15.0/parallelism})}) $2$ Nvidia A100 80GB GPUs requiring thus a total of $8$ Nvidia A100 80GB GPUs to run an experiment ($2$ GPUs $\times 4$ LLM instances). For \textbf{Flan-T5 80M} and \textbf{Flan-T5 3B}, we respectively use $1$ Nvidia V100 32GB and $4$ Nvidia A100 80GB per LLM instance.

In total, to conduct experiments and ablations we use $160$ GPU.hours on the Nvidia V100 32G and $18880$ GPU.hours on Nvidia A100 80GB.

\section{finetuning details}
\label{appendix:finetuning}
\subsection{PPO finetuning details} \label{appendix:ppo_finetuning}
We reused PPO's hyperparameters from \citet{ramamurthy-etal-2022-is} and did not perform any further tuning (see Table \ref{tab:PPO_hp}). We used an Adam \citep{kingma-etal-2014-adam} optimizer with the hyperparameters listed in Table \ref{tab:Adam_hp}). For additional heads, we used MLPs with $3$ hidden layers of $1024$ units with Sigmoid activation.

\begin{table}[H]
    \caption{PPO hyperparameters}
    \label{tab:PPO_hp}
    \centering
    \begin{tabular}{ll}
        \toprule
        \textbf{Variable}  & \textbf{Value} \\
        \midrule
        Number of transitions collected between two updates & $1280$ ($32$ environments $\times40$  steps in each environment) \\
        Number of epochs per update & $4$ \\
        Batch size & $64$ \\
        Entropy loss coefficient & $0.01$ \\
        Value function loss coefficient & $0.5$ \\
        Discount factor & $0.99$ \\
        lr & $1 \times 10^{-6}$ \\
        $\lambda$ factor of the Generalized Advantage Estimator & $0.99$ \\
        Clipping parameter $\epsilon$ & $0.2$ \\
        Maximum gradient norm & $0.5$ \\
        \bottomrule
    \end{tabular}
\end{table}

\begin{table}[H]
    \caption{Adam hyperparameters}
    \label{tab:Adam_hp}
    \centering
    \begin{tabular}{ll}
        \toprule
        \textbf{Variable}  & \textbf{Value} \\
        \midrule
        Learning rate & $1 \times 10^{-6}$ \\
        $\epsilon$ & $1 \times 10^{-5}$ \\
        $\beta_1$ & $0.9$ \\
        $\beta_2$ & $0.999$ \\
        \bottomrule
    \end{tabular}
\end{table}



\subsection{Behavioral Cloning}
\label{appendix:bc}
In Section \ref{sec:results_q4}, we show how grounding using RL differs from BC. For this, we finetune \textbf{Flan-T5 780M} on $400.000$ transitions collected on the \textit{Go To $<$object$>$} task. As indicated in Table \ref{tab:complement_tests_Q3}, GFlan-T5 obtains a $0.81$ success rate on the $1000$ test episodes of the \textit{Go To $<$object$>$} task. Hence by finetuning Flan-T5 to imitate GFlan-T5, one could expect an on-par performance (or worse, but not better). We therefore use GFlan-T5 to collect $400.000$ transitions and finetune Flan-T5 using them. However, the stochasticity in the GFlan-T5 policy leads to deceptive transitions in the dataset (potentially harmful for BC). We thus also assess whether using optimal transitions to finetune Flan-T5 leads to better results than GFlan-T5. To collect optimal trajectories, we use the bot provided by BabyAI and also gather $400.000$ transitions on the \textit{Go To $<$object$>$} task.

For finetuning, we use Causal Language Modeling with the same prompt as the one given to our LLM agents in Section \ref{sec:experiments} as input and the performed action as label. We use the same learning rate as the one used by \citet{rae_scaling_2022} to generate Flan-T5 (i.e. $5\times10^{-4}$) and perform a single epoch on the $400.000$ examples.

\section{Confidence interval}
\label{appendix:confidence_interval}
In sections \ref{sec:results_q2} and \ref{sec:results_q3}, we perform several generalization tests. For each test we report the success rate over $2$ seeds tested on $1000$ episodes each. In the following, we explain how we get the $99\%$ confidence interval.

\subsection{Confidence intervals for GFlan-T5, Flan-T5 and DRRN} 
We model the success of an agent, trained with the seed $i$, on a task (i.e. episode with its associated task) using a Bernoulli variable $X^i \sim \mathcal{B}(p_i)$, with $p_i$ the probability of success of the agent . The number of successes after doing $n$ episodes is the random variable $Y^i_n = \sum_{k=0}^n X^i_k$ which follows a binomial law $\mathcal{B}(n, p_i)$. If $n$ is large enough, the binomial distribution can be approximated by a normal distribution\footnote{using the Berry-Essen theorem, the approximation is good enough if $n>9\frac{p(1-p)}{p}$ and $n>9\frac{p}{p(1-p)}$}. Thus we have
\begin{equation}
\label{equ:distribs}
    \begin{cases}
    p_i & \sim \mathcal{N}(p, \tau^2) \\
    Y^i_n | p_i & \sim \mathcal{N}(n \, p_i, n \, p_i(1-p_i))
    \end{cases}
\end{equation}
where $p$ is the mean success rate and $\tau$ the variance.

Moreover, one property of normal random variables is that if 
\begin{equation}
    \begin{cases}
        V & \sim \mathcal{N}(V_0, \Sigma_V) \\
   U | V & \sim \mathcal{N}(U_0+XV, \Sigma_{U|V})
    \end{cases}
\end{equation}
for any $X$,then 
\begin{equation}
    \begin{pmatrix}
    U \\
    V
    \end{pmatrix} \sim 
    \mathcal{N}(\begin{pmatrix}
    U_0+XV_0\\
    V_0
    \end{pmatrix}, 
    \begin{pmatrix}
    X\Sigma_VX^T+ \Sigma_{U|V} & X\Sigma_V\\
    \Sigma_VX^T & \Sigma_V
    \end{pmatrix})
\end{equation}

Hence we obtain $U \sim \mathcal{N}(U_0+XV_0, X\Sigma_VX^T+ \Sigma_{U|V})$.

By identification with Equation \ref{equ:distribs}, we have 
\begin{equation}
    Y^i_n \sim \mathcal{N}(np, (n\tau)^2+n \, p_i(1-p_i))
\end{equation}

We can rewrite it using the random variable $SR_i$ the success rate of the the agent (trained with seed $i$) during the test time (over $n$ trajectories).

\begin{equation}
    SR_i=\frac{Y^i_n}{n} \sim \mathcal{N}(p, \tau^2+\frac{p_i(1-p_i)}{n})
\end{equation}

Because $\frac{p_i(1-p_i)}{n} \leq \frac{0.25}{n} \xrightarrow[n\to\infty]{} 0$ and $n$ is large, we can neglect this term with respect to $\tau$ in equation above (we verify at the end that we rightfully neglected it) and obtain:

\begin{equation}
    SR_i \sim \mathcal{N}(p, \tau^2)
\end{equation}

Using the maximum likelihood estimation for normal random variables, we get with a $99\%$ confidence interval:
\begin{equation}
    \hat{p} \pm \; 2.58 \frac{\hat{\tau}}{\sqrt{s}} 
\end{equation}
\begin{equation}
    \begin{cases}
        \hat{p} & = \frac{1}{s}\sum_{i=1}^s SR_i \\
        \hat{\tau}^2 & = \frac{1}{(s-1)}\sum_{i=1}^s(\hat{p}-SR_i)^2\\
    \end{cases}
\end{equation}
with $s$ the number of seeds used, $\hat{p}$ the estimator for $p$ and $\hat{\tau}^2$ the unbiased sample variance. 

\subsection{Confidence intervals for random agents}
As previously mentioned, we model the success of an agent using a Bernoulli variable $X \sim \mathcal{B}(p)$, with $p$ the probability of success. The measured success rate after doing $n$ episodes is the random variable $SR_n = \frac{1}{n}\sum_{k=0}^n X_k$ which also follows Bernoulli's law $\mathcal{B}(p)$. Following Hoffending's inequality, we have:
\begin{equation}
    \mathbb{P}(|SR_n - p|>\varepsilon)<2\exp{(-2n\varepsilon^2)=\delta}
\end{equation}
with $\delta$ the error.

Thus if we use $n=1000$ episodes to measure the success rate and we want a confidence of $99\%$ ($\delta=0.01$) with $\varepsilon=\sqrt{|\frac{1}{2n}\ln{\frac{\delta}{2}}|}$, we get $\varepsilon=0.05$. 

\section{Word substitutions for generalization tests}
\label{appendix:generalization_dicts}
For the generalization tests given in the sections \ref{sec:results_q2}, \ref{sec:results_q3}, and \ref{appendix:generalization_test}, we use the dictionaries given below to substitute some words by others.
\subsection{Out of vocabulary}
\label{appendix:out_voc}
To generate descriptions with out-of-vocabulary nouns and adjectives, we modify the prompt by substituting words as per Table \ref{tab:n_adj_mod}.

\begin{table}[htpb]
    \caption{Out-of-vocabulary substitutions for Nouns and adjectives}
    \label{tab:n_adj_mod}
    \centering
    \begin{tabular}{ll}
        \toprule
        \textbf{Original Word}  & \textbf{New Word} \\
        \midrule
        key & chair \\
        ball & table \\
        box & car \\
        \hline
        red & vermillion \\
        green & jade \\
        blue & cyan \\
        purple & violet \\
        yellow & golden \\
        grey & silver \\
        \bottomrule
    \end{tabular}
\end{table}

\subsection{Invented words}
\label{appendix:invented_voc}
Similarly to Section \ref{appendix:out_voc}, we apply the substitutions indicated in Table  \ref{tab:invented_mod}.

\begin{table}[htpb]
    \caption{Invented substitutions for Nouns and adjectives}
    \label{tab:invented_mod}
    \centering
    \begin{tabular}{ll}
        \toprule
        \textbf{Original Word}  & \textbf{New Word} \\
        \midrule
        key & dax \\
        ball & xolo \\
        box & azfe \\
        \hline
        red & faze \\
        green & jatu \\
        blue & croh \\
        purple & vurst \\
        yellow & gakul \\
        grey & sil \\
        \bottomrule
    \end{tabular}
\end{table}

\subsection{Synonym actions}
\label{appendix:synonym_voc}

In~\ref{tab:syn_mod}, we choose the synonym actions to avoid as much as possible to reuse already used words in the finetuning (only "left" and "right" cannot be changed). To verify that Flan-T5-Large considers these words as synonyms we ask it: \textit{"Answer the following yes/no question by reasoning step-by-step. Are $<$original action$>$ and $<$synonym action$>$ synonymous?"}. We retain the synonym only if it considers that this is the case.

\begin{table}[htpb]
    \caption{Synonym actions}
    \label{tab:syn_mod}
    \centering
    \begin{tabular}{ll}
        \toprule
        \textbf{Original Words}  & \textbf{Synonyms} \\
        \midrule
        turn left & rotate left \\
        turn right & rotate right \\
        go forward & move ahead \\
        pick up & take \\
        drop & release \\
        toggle & switch \\
        \bottomrule
    \end{tabular}
\end{table}

\subsection{Translation to French}
\label{appendix:translation}

We give in Table~\ref{tab:fr_translation} the chosen translation for the french environment (the adjectives are given in the feminine form as all the objects are feminine).

\begin{table}[htpb]
    \caption{French translation}
    \label{tab:fr_translation}
    \centering
    \begin{tabular}{ll}
        \toprule
        \textbf{English}  & \textbf{French} \\
        \midrule
        turn left & tourner à gauche \\
        turn right & tourner à droite \\
        go forward & aller tout droit \\
        pick up & prendre \\
        drop & lâcher \\
        toggle & basculer \\
        go to a/the adj n & aller à une/la n adj \\
        steps & pas \\
        You see a $<$\textit{object}$>$ $<$\textit{location}$>$ & Tu vois une $<$objet$>$$<$location$>$ \\
        You see a(n) \textit{open/closed} door $<$\textit{location}$>$ & Tu vois une porte \textit{ouverte/fermée} $<$location$>$  \\
        You carry a $<$\textit{object}$>$ & Tu portes un $<$objet$>$ \\
        key & clef \\
        ball & balle \\
        box & boîte \\
        red & rouge \\
        green & verte \\
        blue & bleue \\
        purple & violette \\
        yellow & jaune \\
        grey & grise \\
        \bottomrule
    \end{tabular}
\end{table}

\end{document}